\documentclass[11pt]{article}

\usepackage[margin=1in]{geometry}
\usepackage{times}
\usepackage{amsmath,amssymb,amsfonts,amsthm,mathtools}
\usepackage{bbm,bm,booktabs,graphicx,microtype,xcolor,tabularx}
\usepackage{placeins}
\usepackage{tikz}
\usetikzlibrary{arrows.meta,positioning}
\usepackage[hyphens]{url}
\usepackage[colorlinks=true,citecolor=blue!60!black,linkcolor=black,urlcolor=blue!60!black]{hyperref}
\usepackage[numbers,sort&compress]{natbib}

\newtheorem{theorem}{Theorem}
\newtheorem{proposition}{Proposition}
\newtheorem{lemma}{Lemma}
\newtheorem{corollary}{Corollary}
\newtheorem{remark}{Remark}

\DeclareMathOperator{\Law}{Law}
\DeclareMathOperator{\Var}{Var}
\DeclareMathOperator{\Cov}{Cov}
\newcommand{\R}{\mathbb R}
\newcommand{\Pbb}{\mathbb P}
\newcommand{\E}{\mathbb E}
\newcommand{\bS}{\bm S}
\newcommand{\bs}{\bm s}
\newcommand{\cP}{\mathcal P}
\newcommand{\cH}{\mathcal H}

\newcommand{\Ujoint}{\mathsf U_{\mathrm{joint}}}
\newcommand{\Umarg}{\mathsf U_{\mathrm{marg}}}
\newcommand{\Useq}{\mathsf U_{\mathrm{seq}}}
\newcommand{\Uworst}{\mathsf U_{\mathrm{worst}}}
\newcommand{\indep}{\perp\!\!\!\perp}

\begin{document}

\title{Multi-Sensitive Distributional Fairness:\\ Joint Wasserstein Audits and Shapley Allocation}
\author{Agathe {Fernandes Machado}, Charlotte Jamotton \& Arthur Charpentier}
\date{}
\maketitle

\begin{abstract}
Fairness audits with several sensitive attributes are often performed one coordinate at a time. This can miss purely intersectional disparities: a score may be independent of each sensitive coordinate while depending on their joint profile. We study distributional demographic parity for a real-valued score using one-dimensional Wasserstein geometry. We distinguish two complementary constructions. Conditional and sequential audits characterize joint parity through their zero sets, but their positive values do not generally decompose a fixed joint Wasserstein disparity. By contrast, full-cell conditional quantile functions define a Hilbert-valued random variable whose variance is the joint disparity. Conditional expectations of this quantile process induce a monotone information game: every reveal order yields an exact martingale decomposition of the same joint disparity, while Shapley values provide an order-free attribution across sensitive attributes. Under independent sensitive attributes, the game reduces to a Hoeffding--ANOVA decomposition and Shapley splits each interaction energy equally among its members; under dependence, orthogonality is lost and explicit formulas reveal suppression and redundancy effects that marginal audits can hide. We give fixed-grid estimators and permutation calibration, and illustrate the framework on synthetic examples and \texttt{ACSIncome}. The resulting framework separates three questions that are often conflated: whether joint distributional parity holds, how a fixed joint disparity decomposes along an information path, and how that disparity can be attributed across sensitive coordinates without privileging an ordering.
\end{abstract}

\section{Introduction}
Fairness audits with several sensitive attributes commonly assess distributional demographic parity separately for sex, race, age, or other coordinates $S_j$ \citep{abraham2020fairnessclusteringmultiplesensitive,jiang2022generalized,hu2024sequential}. This is statistically convenient but incomplete: $Z\indep S_j$ for every $j$ does not imply $Z\indep\bS$. The gap is the distributional analogue of the intersectional and fairness-gerrymandering problem studied in subgroup fairness \citep{kearns2018preventing,hebertjohnson2018multicalibration,foulds2020intersectional,morina2019auditing,lee2025machinelearningmultitypeprotected}.

Given a scalar score $Z$ and categorical sensitive profile $\bS$, we ask two distinct questions: whether the score distribution differs across full intersectional groups, and, conditional on a positive joint disparity, how that fixed total can be allocated across the coordinates of $\bS$. These are deliberately separated as a \emph{certification} problem and a \emph{decomposition} problem. The second question is not answered by summing ordinary conditional Wasserstein audits. In one dimension, the quantile map embeds cell laws in $L^2(0,1)$, but quantiles do not commute with mixture. Consequently, the quantile of $\Law(Z\mid S_A)$ generally differs from the conditional mean of the full-cell quantile process. This distinction is the organizing principle of the paper: conditional laws are used to ask whether parity holds, whereas projections of the full-cell quantile process are used to decompose a fixed joint disparity.

Let $\mathcal Q$ denote the random full-cell quantile function. Its Hilbert-space variance is exactly the joint Wasserstein barycentric disparity, while $M_A=\E(\mathcal Q\mid S_A)$ is its projection onto the information generated by a coalition $A$. The resulting value function is a classical variance-projection game applied to a distribution-valued object. Our contribution is not a new Shapley identity: it is to identify the quantile process as the object that makes this game exact for a joint $W_2^2$ disparity, and to characterize precisely how that construction differs from conditional-law auditing. This positioning connects the paper to global sensitivity analysis \citep{owen2014sobol,song2016shapley,owen2017dependent,gamboa2013multivariate,fort2021global} and to Wasserstein fairness and interpretability \citep{dubey2019frechet,gordaliza2019obtaining,chzhen2020fair,miroshnikov2022wasserstein,benesse2024fairness,mase2021cohort}.

The distinction is interpretive as well as algebraic. A Shapley share here measures how much information a coordinate reveals about the full-cell quantile process, averaged over reveal contexts. It is a descriptive allocation of a chosen barycentric disparity, not a causal effect, legal responsibility, or the disparity removed by intervening on that attribute. Dependence and coding choices can therefore change the shares.

\paragraph{Contributions.}
First, marginal audits and every complete conditional-sequential audit are shown to have different roles but the appropriate zero sets: the former certify only coordinate-wise parity, while the latter vanishes exactly under joint parity. Second, we isolate the mixture--barycenter obstruction and show that conditional and barycentric increments necessarily coincide at the terminal reveal step; the resulting sequential discrepancy is exact but generally signed. Third, the projection game gives nonnegative path decompositions of a fixed $\Ujoint$ and an order-free Shapley allocation, with an ANOVA interpretation under independent coordinates. Fourth, we give fixed-grid estimation, permutation calibration at the degenerate joint null, a marginal-versus-joint power study, and an \texttt{ACSIncome} illustration. Proofs and extended diagnostics are deferred to the appendix.

\section{Joint audits and the barycentric target} \label{sec:def-audits}
Let $Z\in\R$ be a scalar score and let $\bS=(S_1,\ldots,S_d) \in \mathcal{S}$ be categorical with finite support. We assume $Z$ has a finite second moment in every positive-probability full cell of $\mathcal{S}$. Write $[d]=\{1,\ldots,d\}$, $S_A=(S_j)_{j\in A}$, and $\mathcal S_A^+=\{a:\Pbb(S_A=a)>0\}$. For a full cell $\bs\in\mathcal S_{[d]}^+$, write $p_{\bs}=\Pbb(\bS=\bs)$, $\nu_{\bs}=\Law(Z\mid\bS=\bs)$ and $Q_{\bs}$ for its quantile function, and define the random Hilbert-valued quantile process $\mathcal Q(\omega)=Q_{\bS(\omega)}$. Since
\[
W_2^2(\mu,\nu)=\int_0^1\{Q_\mu(u)-Q_\nu(u)\}^2\,du,
\]
the joint Wasserstein disparity is
\begin{equation}\label{eq:ujoint}
\Ujoint(Z)=\inf_{\nu\in\cP_2(\R)}\sum_{\bs}p_{\bs}W_2^2(\nu_{\bs},\nu)
=\E\|\mathcal Q-\bar Q\|_{\cH}^2,
\end{equation}
where $\cH=L^2(0,1)$ and $\bar Q=\E(\mathcal Q)$ is the quantile function of the one-dimensional Wasserstein barycenter $\bar\nu$ \citep{agueh2011barycenters,santambrogio2015optimal}. Since the full support is finite and every positive-probability cell has a finite second moment, $\E\|\mathcal Q\|_{\cH}^2=\E Z^2<\infty$, so all Hilbert-valued conditional expectations below are well defined.

For a single coordinate $S_j$, let $\mathsf U_j$ be the analogous Wasserstein disparity across the actual conditional laws of $\Law(Z\mid S_j=s)$ and define $\Umarg=\sum_j\mathsf U_j$. For disjoint $A,B\subseteq[d]$, let $\mathsf U_{A\mid B}$ denote the $\Pbb(S_B=b)$-weighted average, over $b\in\mathcal S_B^+$, of the Wasserstein disparity across the conditional laws of $\Law(Z\mid S_A=a,S_B=b)$, using within each stratum the conditional weights $\Pbb(S_A=a\mid S_B=b)$ and only compatible positive-probability cells. Thus $\mathsf U_{A\mid B}$ compares the distribution of $Z$ across the levels of $S_A$ \emph{after fixing} $S_B$. Following \cite{hu2024sequential}, for an ordering $\pi$ over the sensitive attributes, the complete conditional-sequential audit is
\[
\Useq^\pi=\sum_{k=1}^d\mathsf U_{\{\pi(k)\}\mid A_{k-1}^\pi},
\qquad 
\begin{cases}
A_k^\pi=\{\pi(1),\ldots,\pi(k)\},\\A_0^\pi=\varnothing.
\end{cases}
\]
For example, with $d=3$ sensitive attributes and a complete ordering $\pi=(2,1,3)$, $\Useq^\pi = \mathsf U_2 + \mathsf U_{1\mid 2} + \mathsf U_{3\mid 1,2}$.

\begin{proposition}[Zero sets]\label{prop:zeros}
$\Umarg=0$ iff $Z\indep S_j$ for every $j$. Moreover, for every complete ordering $\pi$,
\[
\Ujoint=0\quad\Longleftrightarrow\quad Z\indep\bS
\quad\Longleftrightarrow\quad \Useq^\pi=0.
\]
Thus marginal independence of the score from every coordinate does not certify independence from the joint sensitive profile.
\end{proposition}
This proposition is deliberately a statement about \emph{zero sets}: away from zero, $\Useq^\pi$ and $\Ujoint$ need not agree numerically and neither is presented as a rescaling of the other. The proof uses only that $W_2$ is separating. A minimal counterexample is $S_1,S_2\stackrel{\rm iid}{\sim}{\rm Bernoulli}(1/2)$ and $Z=\mathbf1\{S_1=S_2\}$. Conditionally on either $S_1$ or $S_2$ alone, $Z$ remains $\mathrm{Bernoulli}(1/2)$, so every marginal audit vanishes; given the pair, however, $Z$ is deterministic. This is the simplest form of a disparity carried purely by an interaction.

\paragraph{Mixture quantiles versus barycentric projections.}
For a coalition $A$ and value $a$, define the two functions
\begin{equation*}\label{eq:mix-vs-bary}
\begin{aligned}
Q_A^{\rm mix}(a)&=Q_{\Law(Z\mid S_A=a)},\
&
\widetilde Q_A(a)&=\E(\mathcal Q\mid S_A=a).
\end{aligned}
\end{equation*}
The first is the quantile of an actual conditional \emph{mixture} of full-cell laws; the second is the quantile of their Wasserstein \emph{barycenter}. Equivalently, the two operations are ``mix the laws, then take a quantile'' versus ``take the full-cell quantiles, then average them.'' In general
\[
Q_A^{\rm mix}(a)\neq \widetilde Q_A(a).
\]
This single distinction separates certification from decomposition throughout the paper: conditional audits are constructed from $Q_A^{\rm mix}$, whereas the additive game below is constructed from $\widetilde Q_A$. The two objects answer different questions and should not be read as competing estimators of the same intermediate quantity.

To see why this matters, let $S_1,S_2$ be independent uniform binary variables and let the deterministic full-cell scores be $(z_{00},z_{01},z_{10},z_{11})=(0,10,4,6)$ (see Appendix~\ref{app:toy-four-cell}). The full-cell barycenter is $\delta_5$, so $\Ujoint=13$ since $S_1\indep S_2$. For the order $1\to2$, the second conditional term ($\mathsf U_{2\mid 1}$) equals $13$, but the first ($\mathsf U_{1}$) compares the mixtures $(\delta_0+\delta_{10})/2$ and $(\delta_4+\delta_6)/2$ and equals $4$. Hence $\Useq^{(1,2)}=17\neq\Ujoint$. The complete sequential-conditional audit has the right zero set, but it is not an additive decomposition of~\eqref{eq:ujoint}.

\paragraph{Why this is specific to $W_2$.}
The obstruction is geometric, not universal. If laws are embedded by their CDFs in an $L^2(dt)$ Cram\'er geometry, then mixture is linear:
$\E\{F_{Z\mid\bS}(t)\mid S_A\}=F_{Z\mid S_A}(t)$.
The associated projection game therefore coincides with ordinary conditional mixing and obeys an exact chain rule. Quantile/$W_2$ geometry instead keeps distances on the scale of the score and yields the barycentric interpretation above, at the price that mixture and projection need not coincide. Appendix~\ref{app:w2-vs-cdf} records the comparison and a counterexample clarifying the resulting null-player interpretation.

For concentration reporting, define the worst-cell distance to the same barycentric reference,
\[
\Uworst=\max_{\bs:p_{\bs}>0}W_2^2(\nu_{\bs},\bar\nu).
\]
Thus $\Ujoint$ is the weighted average of the full-cell squared distances to $\bar\nu$, whereas $\Uworst$ identifies the most discrepant retained cell. If $p_{\min}$ is the smallest retained cell probability, then $p_{\min}\Uworst\leq\Ujoint\leq\Uworst$; both quantities use the same fixed barycentric reference, not different optimizing centers. Further discussion is given in Appendix~\ref{app:worst-bound}.

\section{The barycentric information game}
\label{sec:barycentric-game}

The conditional audits of Section~\ref{sec:def-audits} are defined from actual conditional laws. They are therefore natural objects for testing distributional parity, but they do not in general decompose the joint Wasserstein disparity. We now introduce a different construction whose purpose is precisely to decompose that fixed joint total $\Ujoint$.

For a coalition $A\subseteq[d]$, define
\[
M_A
=
\E(\mathcal Q\mid S_A),
\qquad
M_\varnothing=\E(\mathcal Q),
\]
and
\begin{equation}
\label{eq:value-game}
\mathcal V(A)
=
\E\|M_A-M_\varnothing\|_{\cH}^2,
\qquad
\cH=L^2(0,1).
\end{equation}
Since a convex combination of one-dimensional quantile functions is again a quantile function, $M_A(a)$ is the quantile function of the Wasserstein barycenter of the full-cell laws compatible with $S_A=a$.

The interpretation of $\mathcal V(A)$ is therefore analogous to explained variance: it measures how much of the variation of the full-cell quantile process $\mathcal Q$ is revealed by the coalition $S_A$. Importantly, $\mathcal V(A)$ is not, in general, the Wasserstein disparity computed from the actual conditional laws $\Law(Z\mid S_A=a)$. The former averages full-cell quantile functions, whereas the latter first mixes the corresponding probability laws.

\begin{theorem}[Barycentric projection game]
\label{thm:main}
For $A\subseteq B\subseteq[d]$,
\[
\mathcal V(B)-\mathcal V(A)
=
\E\|M_B-M_A\|_{\cH}^2
\geq0.
\]
Moreover,
\[
\mathcal V(\varnothing)=0,
\qquad
\mathcal V([d])=\Ujoint.
\]
Hence, for every ordering $\pi$ of $[d]$, with
\[
A_k^\pi=\{\pi(1),\ldots,\pi(k)\},
\qquad A_0^\pi=\varnothing,
\]
we have
\[
\Ujoint
=
\sum_{k=1}^d
\left\{
\mathcal V(A_k^\pi)-\mathcal V(A_{k-1}^\pi)
\right\}.
\]
\end{theorem}

The theorem is a Hilbert-space Pythagorean identity. Indeed, $M_A$ is the orthogonal projection of $\mathcal Q$ onto the $S_A$-measurable subspace of $L^2(\Omega;\cH)$; when $A\subseteq B$, these subspaces are nested. It follows that the extra projection revealed by $B$ is orthogonal to what was already revealed by $A$. Hence every monotone path through the coalition lattice decomposes the same fixed total $\Ujoint$.

\subsection{Barycentric increments and conditional audits}
\label{sec:increments}

For $j\notin A$, define the barycentric increment
\begin{equation}
\label{eq:H-increment}
H_{j\mid A}
:=
\mathcal V(A\cup\{j\})-\mathcal V(A)
=
\E\|M_{A\cup\{j\}}-M_A\|_{\cH}^2.
\end{equation}
Thus $H_{j\mid A}$ measures the additional variation in the full-cell quantile process explained by revealing $S_j$ after $S_A$ is already known. By Theorem~\ref{thm:main}, these increments are nonnegative and, along every ordering $\pi$,
\begin{equation}
\label{eq:H-telescope}
\sum_{k=1}^d
H_{\pi(k)\mid A_{k-1}^\pi}
=
\Ujoint.
\end{equation}

The notation deliberately parallels the conditional audit
$\mathsf U_{\{j\}\mid A}$ introduced in Section~\ref{sec:def-audits}, but the two quantities need not agree. The audit is computed from actual conditional laws, whereas $H_{j\mid A}$ is a projection increment of the full-cell quantile process. Their discrepancy can arise whenever sensitive coordinates remain unrevealed: the audit marginalizes them by mixing probability laws, while the projection game averages their quantile functions.

There is, however, one important point at which the two constructions always coincide.

\begin{proposition}[Terminal-step equivalence]
\label{prop:terminal-equivalence}
Let $j\in[d]$ and $A=[d]\setminus\{j\}$. Then
\[
\mathsf U_{\{j\}\mid A}
=
H_{j\mid A}
=
\mathcal V([d])-\mathcal V(A).
\]
No independence assumption on the sensitive coordinates is required.
\end{proposition}

At the terminal step, conditioning on $S_A$ leaves only $S_j$ unrevealed.
The conditional laws entering the audit are therefore already full-cell laws: there is no remaining sensitive coordinate to marginalize, and the mixture--barycenter discrepancy disappears.

This observation localizes exactly why a complete conditional-sequential audit need not equal the joint disparity. Define
\[
\Delta_k^\pi
=
\mathsf U_{\{\pi(k)\}\mid A_{k-1}^\pi}
-
H_{\pi(k)\mid A_{k-1}^\pi}.
\]

\begin{proposition}[Sequential discrepancy decomposition]
\label{prop:sequential-discrepancy}
For every ordering $\pi$,
\begin{equation}
\label{eq:sequential-discrepancy}
\Useq^\pi-\Ujoint
=
\sum_{k=1}^{d-1}\Delta_k^\pi.
\end{equation}
In particular,
\[
\Delta_d^\pi=0.
\]
Hence the difference between the conditional-sequential audit and the joint Wasserstein disparity is entirely accumulated at nonterminal steps, while sensitive coordinates still remain to be marginalized.
\end{proposition}

For two sensitive coordinates, this becomes especially transparent.

\begin{corollary}[Two sensitive coordinates]
\label{cor:two-sensitive}
For $d=2$,
\[
\mathsf U_{2\mid1}
=
H_{2\mid1}
=
\mathcal V(\{1,2\})-\mathcal V(\{1\}),
\]
and
\[
\mathsf U_{1\mid2}
=
H_{1\mid2}
=
\mathcal V(\{1,2\})-\mathcal V(\{2\}).
\]
Consequently,
\[
\Useq^{(1,2)}
=
\Ujoint+
\left\{
\mathsf U_1-\mathcal V(\{1\})
\right\},
\]
and
\[
\Useq^{(2,1)}
=
\Ujoint+
\left\{
\mathsf U_2-\mathcal V(\{2\})
\right\}.
\]
Thus, with two sensitive coordinates, the entire failure of the conditional-sequential audit to telescope to $\Ujoint$ occurs at the first step.
\end{corollary}

For the four-cell example of Section~\ref{sec:def-audits}, $\mathsf U_1=4$ while $\mathcal V(\{1\})=0$. Corollary~\ref{cor:two-sensitive} therefore gives $\Useq^{(1,2)}=13+(4-0)=17$: the discrepancy observed earlier is entirely generated at the first step.

\begin{remark}[Signed discrepancy and null players]\label{rem:signed-gap}
Neither $\Delta_k^\pi$ nor $\Useq^\pi-\Ujoint$ has a fixed sign, so the sequential audit is in general neither an upper nor a lower bound for the joint barycentric disparity. Moreover, marginal independence $Z\indep S_j$ need not imply $\mathcal V(\{j\})=0$: equality of conditional mixture laws does not force equality of the corresponding Wasserstein barycenters. Thus a positive Shapley share is an information allocation in the quantile-projection game, not a certificate of marginal dependence. Appendix~\ref{app:w2-vs-cdf} gives a two-coordinate counterexample for which $Z\indep S_1$ but $\mathcal V(\{1\})>0$ and $\Useq^{(1,2)}<\Ujoint$.
\end{remark}

\subsection{Order-free allocation}

The path increments in \eqref{eq:H-telescope} provide exact
decompositions of $\Ujoint$, but the amount assigned to a coordinate may depend on when that coordinate is revealed. The Shapley value removes this ordering choice by averaging the corresponding barycentric increment over all reveal contexts.

For $j\in[d]$, define
\begin{equation}
\label{eq:shapley}
\phi_j
=
\sum_{A\subseteq[d]\setminus\{j\}}
\frac{|A|!(d-|A|-1)!}{d!}
H_{j\mid A}.
\end{equation}

Equivalently, if $P_j(\pi)$ denotes the set of coordinates preceding $j$ in a uniformly random permutation $\pi$,
\[
\phi_j
=
\E_\pi
\left[
H_{j\mid P_j(\pi)}
\right].
\]

Since every $H_{j\mid A}$ is nonnegative,
\[
\phi_j\geq0,
\qquad
\sum_{j=1}^d\phi_j=\Ujoint.
\]
Thus Shapley does not define a new fairness target. It averages the \emph{barycentric path increments} in~\eqref{eq:H-telescope}, not the conditional-audit terms, and provides an order-free allocation of the fixed barycentric disparity $\Ujoint$ across sensitive coordinates.

\section{Interactions and dependence}
\label{sec:interactions}

For every coalition,
\begin{equation}\label{eq:scalar-integral-game}
\mathcal V(A)=\int_0^1
\Var\!\left\{\E\bigl[\mathcal Q(u)\mid S_A\bigr]\right\}\,du.
\end{equation}
Thus the game is the integral, over quantile level $u$, of ordinary scalar variance games. This makes explicit what is classical in the construction and what is specific to the Wasserstein audit: the scalar projection/Shapley machinery is standard, while the quantile embedding makes its total equal to $\Ujoint$.

\subsection{Independent sensitive coordinates}
If $S_1,\ldots,S_d$ are mutually independent, define $h_\varnothing=M_\varnothing$ and
$h_A(S_A)=M_A-\sum_{B\subsetneq A}h_B(S_B)$. The Hilbert-valued Hoeffding decomposition gives mutually orthogonal components
\[
\mathcal Q=h_\varnothing+\sum_{\varnothing\neq A\subseteq[d]}h_A(S_A),
\qquad
\mathsf I_A=\E\|h_A(S_A)\|_{\cH}^2,
\]
so that
\begin{equation}\label{eq:anova}
\Ujoint=\sum_{A\neq\varnothing}\mathsf I_A,
\qquad
\mathcal V(C)=\sum_{\varnothing\neq A\subseteq C}\mathsf I_A.
\end{equation}
Consequently,
\begin{equation}\label{eq:H-anova}
H_{j\mid A}=\sum_{\substack{B\subseteq A\cup\{j\}\\j\in B}}\mathsf I_B,
\qquad
\phi_j=\sum_{B\ni j}\frac{\mathsf I_B}{|B|}.
\end{equation}
Here $\mathsf I_B$ is the orthogonal energy of the quantile-process interaction indexed by $B$. A reveal path assigns that energy to the last member of $B$ revealed, while Shapley averages over orders and shares $\mathsf I_B$ equally among its members \citep{owen2014sobol}. This is an ANOVA statement about the quantile process, not a claim that a lower-order conditional-law disparity is present or absent.

For a pure highest-order quantile interaction,
\begin{equation}\label{eq:pure-interaction}
\mathcal Q(u)=Q_0(u)+a\prod_{j=1}^dS_j,
\qquad S_j\stackrel{\rm iid}{\sim}\mathrm{Unif}\{-1,1\},
\end{equation}
all strict coalition values vanish, $\Ujoint=a^2$, every path assigns the total to the last revealed coordinate, and $\phi_j=a^2/d$. Thus a joint disparity can be invisible to every strict lower-order coalition. Appendix~\ref{app:theory} gives the full functional-ANOVA argument.

\subsection{Dependent sensitive coordinates}
Under dependence, $\mathcal V$, $H_{j\mid A}$ and $\phi_j$ remain well defined and nonnegative, but the orthogonal ANOVA interpretation need not hold \citep{owen2017dependent,iooss2019shapley}. Reveal context can matter through redundancy, complementarity, masking, or interactions. We use the descriptive envelope
\[
H_j^- = \min_{A\not\ni j}H_{j\mid A},
\qquad
H_j^+ = \max_{A\not\ni j}H_{j\mid A},
\]
for which $H_j^-\le\phi_j\le H_j^+$ because $\phi_j$ is a weighted average of the increments. This range is a diagnostic, not an intrinsic dependence measure: it can be wide even for independent coordinates when interactions are present.

A Gaussian location model gives a closed-form illustration. Although it lies outside the categorical setup used for the audit estimand, it isolates the effect of dependence without numerical approximation. If $\bS\sim N(0,\Sigma)$ and $Z=\beta^\top\bS+\varepsilon$ with independent Gaussian noise, then
\[
\begin{aligned}
\mathcal V(A)&=\Var\{\E(Z\mid S_A)\}\\
&=c_A^\top\Sigma_{AA}^{-1}c_A,
\qquad c_A=\Cov(S_A,Z).
\end{aligned}
\]
For standardized $(S_1,S_2)$ with correlation $\rho$,
\begin{equation}\label{eq:gaussian}
\mathcal V(\{2\})=(\beta_2+\rho\beta_1)^2,
\qquad
H_{2\mid1}=\beta_2^2(1-\rho^2).
\end{equation}
If $\beta_1=-\beta_2$ and $\rho>0$, then
$H_{2\mid1}/\mathcal V(\{2\})=(1+\rho)/(1-\rho)$: marginal cancellation can make a coordinate weak when revealed alone but informative once another coordinate is known. The corresponding figure and further formulas are in Appendix~\ref{app:dependent-gaussian}.

\section{Estimation and calibration}

Statistical inference for Wasserstein variation around barycenters, including central-limit and bootstrap results, has been studied in related settings by \citet{delbarrio2019wasserstein}. Here we deliberately separate inference for a fixed quantile grid from numerical approximation of the integrated $W_2^2$ target. This yields a finite-dimensional inferential problem, allowing regular and degenerate regimes of the multi-cell audit to be distinguished without simultaneously imposing the tail regularity needed for increasingly fine quantile grids.

Fix $0<u_1<\cdots<u_K<1$. For a full cell
$s\in\mathcal S$, let $\widehat Q_s(u_k)$ denote the empirical quantiles of
$\Law(Z\mid \bS=s)$, let $\widehat p_s=n_s/n$, and define
\[
\widehat p_A(a)=\sum_{s:s_A=a}\widehat p_s.
\]
Then
\[
\widehat M_A(a;u_k)=
\sum_{s:s_A=a}
\frac{\widehat p_s}{\widehat p_A(a)}
\widehat Q_s(u_k),
\]
and
\begin{equation}\label{eq:estimator}
\widehat{\mathcal V}_K(A)=
\sum_a\widehat p_A(a)\frac1K\sum_{k=1}^K
\{\widehat M_A(a;u_k)-\widehat M_\varnothing(u_k)\}^2.
\end{equation}
Write $\mathcal V_K(A)$ for the population finite-grid analogue and set $\widehat\Ujoint_K=\widehat{\mathcal V}_K([d])$. Fixed-$K$ inference targets $\mathcal V_K$; approximation of $\mathcal V$ by a grid is a separate deterministic issue treated in Appendix~\ref{app:grid-approx}.

\begin{proposition}[Fixed-grid regimes]\label{prop:grid}\label{prop:consistency}
Assume a fixed finite cell structure with $p_s>0$, i.i.d. observations, and cell CDFs that are continuous and strictly increasing with positive continuous densities at the grid quantiles. Then $\widehat{\mathcal V}_K(A)\xrightarrow{p}\mathcal V_K(A)$ for every fixed $A$. At a regular point where the first-order asymptotic variance is nonzero,
\[
\sqrt n\{\widehat{\mathcal V}_K(A)-\mathcal V_K(A)\}
\Rightarrow N(0,\sigma_{A,K}^2).
\]
Under the exact joint null $Z\indep\bS$,
\begin{equation}\label{eq:quadratic-null-main}
n\widehat\Ujoint_K\Rightarrow
\sum_s p_s\|G_s^q-\bar G\|_K^2,
\end{equation}
where $\bar G=\sum_sp_sG_s^q$ and $G_s^q$ is the Gaussian cell-quantile limit for $s \in \mathcal{S}$.
\end{proposition}
Thus the stochastic scale changes at exact parity: regular alternatives fluctuate on the $n^{-1/2}$ scale, whereas the joint statistic under the null is of order $n^{-1}$. A first-order Wald calibration is therefore inappropriate at parity. Degeneracy can also occur for strict coalitions with $\mathcal V_K(A)=0$; standard first-order intervals for coalition increments should then be interpreted cautiously. The appendix gives the cellwise quantile CLT and delta-method details.

The statistical price of intersectional cells is visible in the quantile limit. At regular grid points,
\begin{equation}\label{eq:rare-cell-clt}
\Cov(G^q_{s,k},G^q_{s,\ell})=
\frac{\min(u_k,u_\ell)-u_ku_\ell}
{p_s f_s\{Q_s(u_k)\}f_s\{Q_s(u_\ell)\}}.
\end{equation}
Small $p_s$ and extreme quantiles with small density inflate variance, so joint auditing cannot uniformly dominate lower-dimensional marginal tests in finite samples.

\begin{proposition}[Permutation calibration]\label{prop:perm}
For fixed held-out scores and sensitive profiles, under $Z\indep\bS$ the score-permutation randomization test is finite-sample valid. With $B$ random permutations, the corrected Monte Carlo p-value
$(1+\#\{T^{(b)}\ge T^{\rm obs}\})/(B+1)$ is super-uniform \citep{phipson2010permutation}.
\end{proposition}
If scores are learned on an independent training sample, the audit scores can be treated as held out. Any score-dependent fitting or selection performed on the audit sample must instead be separated from the audit or rerun inside each randomization replicate.

\paragraph{Retained support.}
A minimum-cell rule based only on $\bS$ defines a retained set $\mathcal R$ and changes the estimand to the renormalized conditional population $\Law(Z,\bS\mid\bS\in\mathcal R)$. It does \emph{not} estimate or impute the disparity of omitted rare intersections. The same retained support is then held fixed across score permutations. Reporting should therefore include retained-cell counts and sensitivity to the threshold and quantile grid. Quantile shrinkage, permutation-sampled Shapley values, and computational details are given in Appendices~\ref{app:stat}--\ref{app:computation}.

\section{Power: when marginal and joint audits differ}
\begin{figure}[t]
\centering
\includegraphics[width=.98\linewidth]{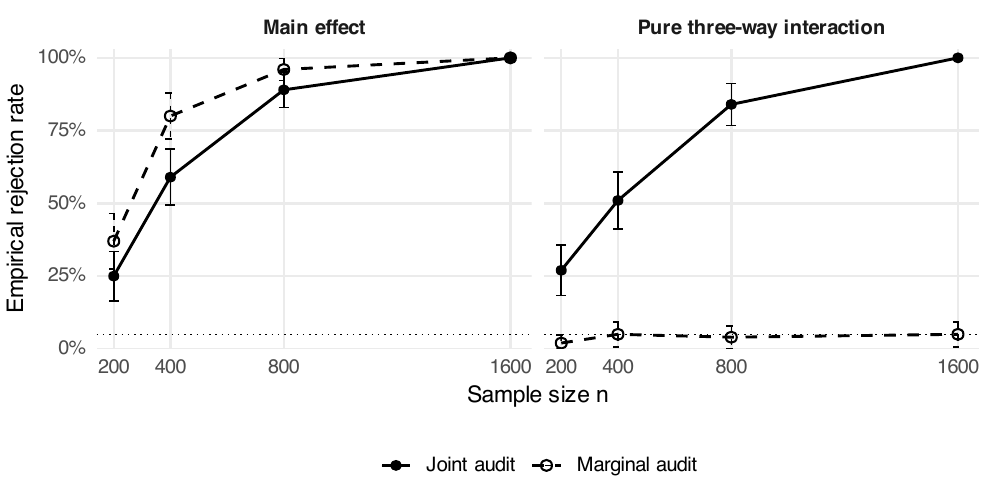}
\caption{Empirical rejection rates at nominal $5\%$ ($\pm1.96$ Monte Carlo s.e.). The marginal statistic is stronger for the main-effect alternative at moderate $n$, whereas it stays near size under the pure interaction while the joint audit gains power.}\label{fig:power}
\end{figure}

Joint parity is stronger than coordinate-wise parity, but estimating full-cell distributions costs precision. We illustrate this complementarity with three independent Rademacher attributes and either a main effect $Z=0.15S_1+\varepsilon$ or a pure three-way interaction $Z=0.15S_1S_2S_3+\varepsilon$. We compare permutation tests based on the joint statistic and on $T_{\rm marg}=\sum_j\widehat{\mathsf U}_j$, calibrating each global statistic by the same score-permutation principle. In the pure interaction design, each one-coordinate conditional score law is unchanged, so a coordinate-wise marginal audit has no population signal even though the full joint profile does. Figure~\ref{fig:power} uses 100 Monte Carlo datasets, 99 permutations, and nine interior quantiles per design point; it is intended as a finite-sample diagnostic rather than a universal power ranking.

For $n=200,400,800,1600$, marginal rejection under the main effect is $37,80,96,100\%$, versus $25,59,89,100\%$ for the joint statistic. Under the pure interaction, marginal rejection remains at $2$--$5\%$, while joint rejection rises from $27\%$ to $51,84,100\%$. Equation~\eqref{eq:rare-cell-clt} explains the variance cost of small full cells; the interaction design shows the complementary blind spot of coordinate-wise audits. Extended null and interaction experiments appear in Appendix~\ref{app:numerics}.

\section{Application: \texttt{ACSIncome}}
We audit held-out regularized-logistic scores for predicting income above USD 50,000, in \texttt{ACSIncome} \citep{ding2021retiring}. The sensitive profile is $(\texttt{SEX},\texttt{RAC1P},\texttt{AGE\_GROUP})$, with age discretized into four groups. For each state, we fit a LASSO-penalized logistic regression using
\texttt{glmnet}, with \texttt{family = "binomial"} and
\texttt{alpha = 1}. The penalty parameter is selected by five-fold
cross-validation using binomial deviance
(\texttt{cv.glmnet(..., nfolds = 5, type.measure = "deviance")}).
The direct predictors are
\texttt{COW}, \texttt{SCHL}, \texttt{MAR}, \texttt{OCCP},
\texttt{POBP}, \texttt{RELP}, and \texttt{WKHP};
the audit attributes \texttt{SEX}, \texttt{RAC1P}, and
\texttt{AGE\_GROUP} are excluded as direct covariates.
We audit the held-out linear predictor
\[
Z=X^\top\widehat\beta,
\]
obtained with \texttt{predict(..., type = "link")}, rather than the predicted probability. Scores are standardized within state before computing Wasserstein disparities. Thus the empirical analysis audits held-out logit scores from a sparse logistic model fitted only on non-sensitive covariates; excluding the sensitive coordinates from the model does not preclude proxy dependence through the remaining predictors.

We use $K=99$ quantiles ($.01,\ldots,.99$), retain full cells with at least 20 held-out observations, and keep that profile-based retained support fixed under 999 score permutations. The resulting estimand is the renormalized retained-support audit; 26--32 cells remain across the ten states. Stratified bootstrap intervals resample scores within retained full cells and are descriptive conditional on the observed profile composition. Full cell-size and state-level diagnostics are in Appendix~\ref{app:acs}.

\begin{figure}[t]
\centering
\includegraphics[width=.68\linewidth]{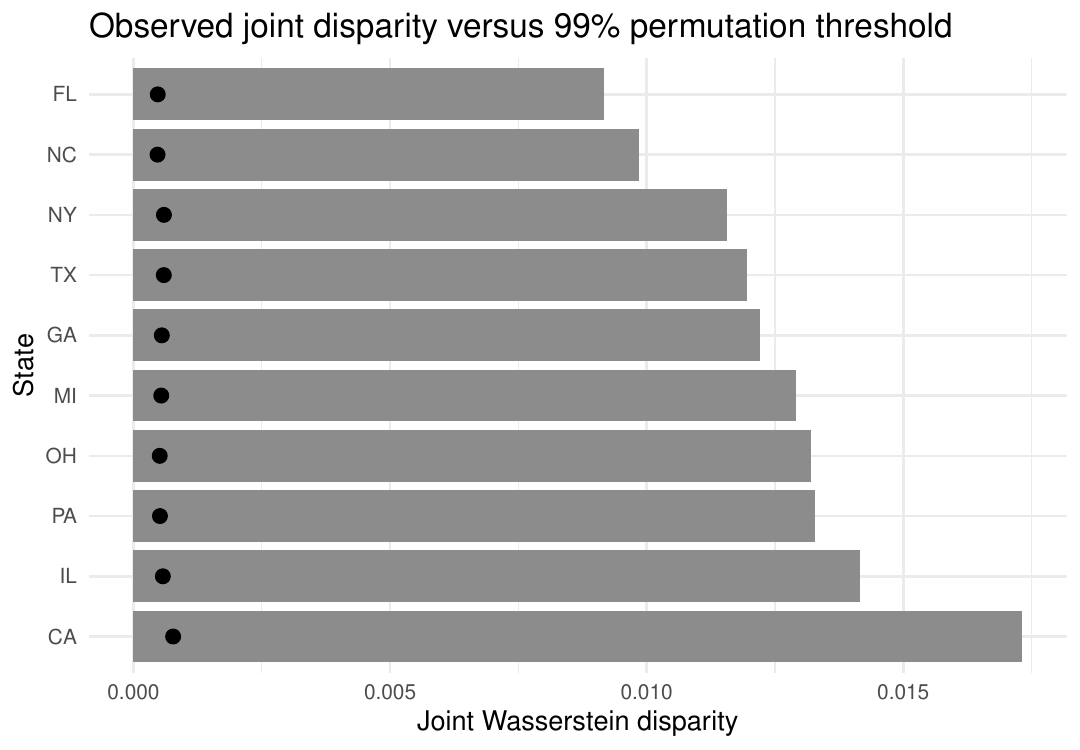}\\[-1mm]
\textbf{(a)} Observed joint disparity vs. 99\% permutation threshold\\[1mm]
\includegraphics[width=.68\linewidth]{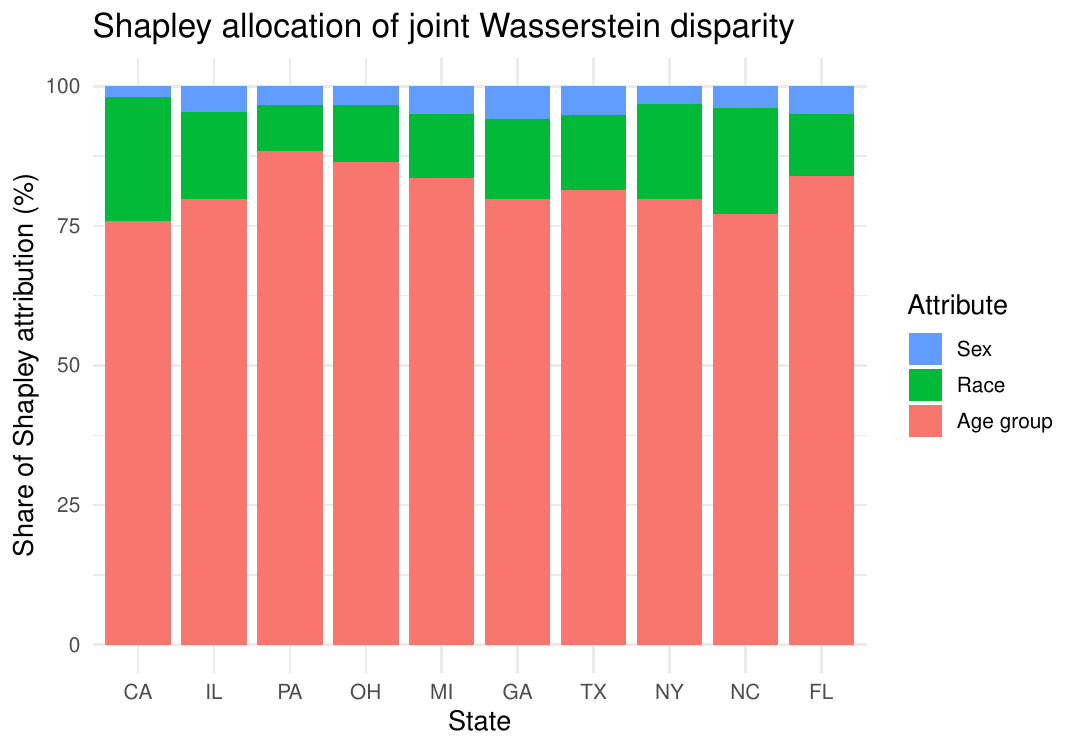}\\[-1mm]
\textbf{(b)} Shapley allocation of the joint disparity
\caption{\texttt{ACSIncome}. The joint statistic is far outside the permutation null in every state. Shapley shares allocate the retained-support barycentric disparity descriptively; they are coding- and grouping-dependent, not causal attributions.}\label{fig:acs}
\end{figure}

Across states, $\widehat\Ujoint$ ranges from $0.0092$ to $0.0173$, every 99\% permutation threshold is below $0.0011$, and every corrected p-value is at its minimum resolution. Since scores are standardized within state, these values are best read as effect-size summaries on a common within-state scale; the permutation p-value addresses the sharper question of exact independence. The ratio $\Uworst/\Ujoint$ ranges from $4.4$ to $8.0$, meaning that the largest retained-cell \emph{squared} $W_2$ distance to the common barycenter is roughly four to eight times the corresponding weighted average. To quantify the dependence of the conditional-sequential audit on the
chosen ordering, we report the relative sequential-order sensitivity
\[
  \text{OrderSens}
  =
    100~
  \frac{
    \max_{\pi}\Useq^\pi-\min_{\pi}\Useq^\pi
  }{
    \Ujoint
  }.
  \]
For the three audited attributes, the maximum and minimum are taken over all $3!=6$ possible orderings. Sequential-order sensitivity is below $1.2\%$ in every state; this empirical stability is not a general invariance result.

Age group receives $75.8$--$88.5\%$ of the Shapley allocation, race $8.2$--$22.1\%$, and sex $2.0$--$5.9\%$. These shares depend on the chosen age bins, race coding, and retained support. In particular, a finer coding can reveal additional quantile-process variation and change the allocation. We therefore interpret the ranking as descriptive for this audit design, not as evidence that intervening on an attribute would remove the corresponding share of disparity. Near the joint null, normalized shares would also be unstable and should not be emphasized.

\FloatBarrier
\section{Discussion}
The paper separates certification from decomposition. Conditional Wasserstein audits compare actual conditional mixture laws and therefore retain the direct law-level interpretation of distributional parity; the projection game answers a different question by decomposing variation in the full-cell quantile process. They coincide at the terminal reveal step but need not coincide earlier, and their discrepancy is signed. Every barycentric reveal path nevertheless telescopes exactly to the fixed joint disparity.

The projection and Shapley identities are inherited from variance-based sensitivity analysis. The Wasserstein-specific contribution is the quantile geometry that makes the variance of $\mathcal Q$ exactly $\Ujoint$. That geometry is also why mixing and projection fail to commute; a linear CDF/Cram\'er embedding would instead obey a chain rule. Hence Shapley shares here are descriptive information allocations, not causal or normative responsibility measures. In particular, a positive share need not certify marginal dependence, and a zero or small share should not be read as a general conditional-parity statement.

Joint auditing is statistically costly when intersections are sparse, and coding or retention choices can change both the estimand and its allocation. Its purpose is not universal power dominance, but certification against higher-order structure invisible to coordinate-wise tests. The current theory is one-dimensional and fixed-cell; multivariate scores and high-dimensional intersectionality require additional regularization. A practical workflow is therefore: define the support, test the joint null, inspect concentrated cellwise discrepancies, diagnose context sensitivity, and only then allocate a positive disparity---\emph{detect first, decompose second, interpret cautiously}.

\bibliographystyle{plainnat}
\bibliography{biblio-aistats}

\clearpage
\appendix

\section{A four-cell toy example}
\label{app:toy-four-cell}

\subsection{Independent sensitive attributes}

This example illustrates, in the simplest possible setting, the distinction between
sequential audits based on actual conditional mixture laws and the barycentric game
used for numerical attribution. Let
\[
S_1,S_2 \overset{\mathrm{ind}}{\sim} \mathrm{Bernoulli}(1/2),
\]
and let the score be deterministic conditional on the sensitive profile:
\[
Z=z_{S_1S_2},
\qquad
(z_{00},z_{01},z_{10},z_{11})=(0,10,4,6).
\]
Thus each of the four sensitive cells has probability $1/4$, and
\[
\begin{array}{c|cc}
 & S_2=0 & S_2=1\\
\hline
S_1=0 & 0 & 10\\
S_1=1 & 4 & 6
\end{array}.
\]

\paragraph{Joint audit.}
Since $Z$ is deterministic within each full cell, the four conditional quantile
functions are constant:
\[
Q_{00}(u)=0,\qquad
Q_{01}(u)=10,\qquad
Q_{10}(u)=4,\qquad
Q_{11}(u)=6,
\qquad 0<u<1.
\]
Their Wasserstein barycenter has constant quantile
\[
\overline Q(u)
=
\frac{0+10+4+6}{4}
=
5.
\]
Hence
\[
U_{\mathrm{joint}}
=
\frac14\left\{
(0-5)^2+(10-5)^2+(4-5)^2+(6-5)^2
\right\}
=
13.
\]

\paragraph{Marginal audit for $S_1$.}
Conditionally on $S_1=0$,
\[
\mathcal L(Z\mid S_1=0)
=
\frac12\delta_0+\frac12\delta_{10},
\]
whereas
\[
\mathcal L(Z\mid S_1=1)
=
\frac12\delta_4+\frac12\delta_6.
\]
Their quantile functions are therefore
\[
Q^{(1)}_0(u)
=
\begin{cases}
0,&0<u<1/2,\\
10,&1/2\leq u<1,
\end{cases}
\qquad
Q^{(1)}_1(u)
=
\begin{cases}
4,&0<u<1/2,\\
6,&1/2\leq u<1.
\end{cases}
\]
The barycenter quantile is
\[
\overline Q^{(1)}(u)
=
\begin{cases}
2,&0<u<1/2,\\
8,&1/2\leq u<1,
\end{cases}
\]
so
\[
U_1
=
\frac12\int_0^1
\left(Q^{(1)}_0(u)-\overline Q^{(1)}(u)\right)^2\,du
+
\frac12\int_0^1
\left(Q^{(1)}_1(u)-\overline Q^{(1)}(u)\right)^2\,du
=
4.
\]

\paragraph{Marginal audit for $S_2$.}
Similarly,
\[
\mathcal L(Z\mid S_2=0)
=
\frac12\delta_0+\frac12\delta_4,
\qquad
\mathcal L(Z\mid S_2=1)
=
\frac12\delta_6+\frac12\delta_{10}.
\]
The corresponding quantile functions are
\[
Q^{(2)}_0(u)
=
\begin{cases}
0,&0<u<1/2,\\
4,&1/2\leq u<1,
\end{cases}
\qquad
Q^{(2)}_1(u)
=
\begin{cases}
6,&0<u<1/2,\\
10,&1/2\leq u<1.
\end{cases}
\]
Their barycenter quantile is $(3,7)$ over the two half-intervals, hence
\[
U_2=9.
\]
In this particular example,
\[
U_1+U_2=4+9=13=U_{\mathrm{joint}},
\]
but this equality is accidental and does not hold in general.

\paragraph{Sequential audits.}
Consider first the order $S_1\to S_2$.
Within the stratum $S_1=0$, revealing $S_2$ distinguishes $\delta_0$ from
$\delta_{10}$, whose barycenter is $\delta_5$, so
\[
U_{2\mid S_1=0}
=
\frac12(0-5)^2+\frac12(10-5)^2
=
25.
\]
Within $S_1=1$, the two conditional laws are $\delta_4$ and $\delta_6$, and
\[
U_{2\mid S_1=1}
=
\frac12(4-5)^2+\frac12(6-5)^2
=
1.
\]
Averaging over the two equally likely strata gives
\[
U_{2\mid 1}
=
\frac12(25)+\frac12(1)
=
13.
\]
Therefore
\[
U_1+U_{2\mid 1}
=
4+13
=
17.
\]

For the reverse order $S_2\to S_1$, within $S_2=0$ the two point masses are
$\delta_0$ and $\delta_4$, while within $S_2=1$ they are $\delta_{10}$ and
$\delta_6$. In both strata the conditional disparity equals $4$, so
\[
U_{1\mid 2}=4,
\]
and consequently
\[
U_2+U_{1\mid 2}
=
9+4
=
13.
\]

Thus
\[
U_1+U_{2\mid1}=17,
\qquad
U_2+U_{1\mid2}=13,
\qquad
U_{\mathrm{joint}}=13.
\]
The sequential audit is therefore order dependent, and its positive value does
not in general telescope to $U_{\mathrm{joint}}$. Its role is instead
zero-set based: vanishing of all sequential terms along an order is equivalent
to joint distributional parity.

\begin{remark}
    Independence of $S_1$ and $S_2$ only says that revealing $S_1$ does not help predict $S_2$. It does not say that $S_1$ is irrelevant to how $S_2$ is associated with the score. $U_{2\mid1}$ compares distributions before mixing over $S_1$, whereas $U_2$ compares them after mixing. Wasserstein distance does not commute with this mixing operation.
\end{remark}

\paragraph{The barycentric game.}
Now define, for $A\subseteq\{1,2\}$,
\[
V(A)
=
\int_0^1
\operatorname{Var}\!\left(
\mathbb E[Q_S(u)\mid S_A]
\right)\,du,
\]
where
\[
Q_S(u)=Q_{Z\mid S_1,S_2}(u).
\]
Here $Q_S(u)=Z$ for every $u$, since the full-cell laws are degenerate.

Clearly,
\[
V(\varnothing)=0.
\]
Conditioning on $S_1$ gives
\[
\mathbb E[Q_S(u)\mid S_1=0]
=
\frac{0+10}{2}
=
5,
\qquad
\mathbb E[Q_S(u)\mid S_1=1]
=
\frac{4+6}{2}
=
5,
\]
hence
\[
V(\{1\})=0.
\]
By contrast, conditioning on $S_2$ gives
\[
\mathbb E[Q_S(u)\mid S_2=0]=2,
\qquad
\mathbb E[Q_S(u)\mid S_2=1]=8,
\]
so
\[
V(\{2\})
=
\frac12(2-5)^2+\frac12(8-5)^2
=
9.
\]
Finally,
\[
V(\{1,2\})
=
\operatorname{Var}(Q_S)
=
13
=
U_{\mathrm{joint}}.
\]
Therefore the complete game is
\[
V(\varnothing)=0,\qquad
V(\{1\})=0,\qquad
V(\{2\})=9,\qquad
V(\{1,2\})=13.
\]

This also makes explicit why the marginal audit and the barycentric game need not agree:
\[
{
U_1=4
\ \text{whereas}\ 
V(\{1\})=0.
}
\]
Indeed,
\[
Q_{\mathcal L(Z\mid S_1)}(u)
\neq
\mathbb E\!\left[
Q_{Z\mid S_1,S_2}(u)\mid S_1
\right]
\]
in general: taking the quantile of a mixture does not commute with averaging
the component quantiles.

\paragraph{Path increments and Shapley allocation.}
For the barycentric game, the path increments do telescope. Along the order
$1\to2$,
\[
H_1
=
V(\{1\})-V(\varnothing)
=
0,
\qquad
H_{2\mid1}
=
V(\{1,2\})-V(\{1\})
=
13,
\]
while along $2\to1$,
\[
H_2
=
V(\{2\})-V(\varnothing)
=
9,
\qquad
H_{1\mid2}
=
V(\{1,2\})-V(\{2\})
=
4.
\]
In both cases,
\[
H_1+H_{2\mid1}
=
H_2+H_{1\mid2}
=
U_{\mathrm{joint}}
=
13.
\]

The Harsanyi dividends are
\[
D_{\{1\}}=0,
\qquad
D_{\{2\}}=9,
\qquad
D_{\{1,2\}}
=
13-0-9
=
4.
\]
Shapley aggregation has a particularly natural interpretation in this example:
it averages the two possible orders in which the sensitive attributes can enter
the barycentric decomposition.

For $S_1$,
\[
\phi_1
=
\frac12
\underbrace{H_1}_{0}
+
\frac12
\underbrace{H_{1\mid2}}_{4}
=
2,
\]
while for $S_2$,
\[
\phi_2
=
\frac12
\underbrace{H_2}_{9}
+
\frac12
\underbrace{H_{2\mid1}}_{13}
=
11.
\]

Equivalently, the $4$ units of interaction ($D_{\{1,2\}}$) are split equally between the two attributes:
\[
\phi_1
=
0+\frac{4}{2}
=
2,
\qquad
\phi_2
=
9+\frac{4}{2}
=
11.
\]

The example can therefore be read in two complementary ways. First, the joint
disparity decomposes into a component attributable to $S_2$ alone and a purely
joint component:
\[
13
=
\underbrace{0}_{S_1\text{ alone}}
+
\underbrace{9}_{S_2\text{ alone}}
+
\underbrace{4}_{\text{interaction}}.
\]

The Shapley allocation therefore assigns
\[
\frac{\phi_1}{U_{\mathrm{joint}}}
=
\frac{2}{13}
\approx 15.4\%
\]
of the joint distributional disparity to $S_1$, and
\[
\frac{\phi_2}{U_{\mathrm{joint}}}
=
\frac{11}{13}
\approx 84.6\%
\]
to $S_2$.
These shares are descriptive allocations of $U_{\mathrm{joint}}$ under the barycentric game and should not be interpreted causally.

\subsection{Correlated sensitive attributes}
\label{app:toy-dependent}

We now keep the same four cell values,
\[
(z_{00},z_{01},z_{10},z_{11})=(0,10,4,6),
\]
but allow the two sensitive attributes to be dependent. Let
\[
S_1,S_2\sim\mathrm{Bernoulli}(1/2),
\]
with joint distribution
\[
\begin{array}{c|cc}
 & S_2=0 & S_2=1\\
\hline
S_1=0 & \dfrac{1+\rho}{4} & \dfrac{1-\rho}{4}\\[2mm]
S_1=1 & \dfrac{1-\rho}{4} & \dfrac{1+\rho}{4}
\end{array},
\qquad -1\leq\rho\leq1.
\]
The marginals remain Bernoulli$(1/2)$ and
\[
\operatorname{Corr}(S_1,S_2)=\rho.
\]
For $|\rho|<1$, all four cells have positive probability. The formulas below
extend continuously to $\rho=\pm1$, where only two cells remain in the support.

As before,
\[
Z=z_{S_1S_2},
\]
so the full-cell conditional laws are point masses and the full-cell quantile
process is constant in $u$.

\paragraph{Joint disparity.}
The Wasserstein barycenter of the full-cell laws is the point mass at
\[
\overline Q
=
\frac{1+\rho}{4}0
+
\frac{1-\rho}{4}10
+
\frac{1-\rho}{4}4
+
\frac{1+\rho}{4}6
=
5-2\rho.
\]
Moreover,
\[
\mathbb E[Z^2]
=
38-20\rho.
\]
Hence
\[
U_{\mathrm{joint}}
=
\operatorname{Var}(Z)
=
38-20\rho-(5-2\rho)^2,
\]
and therefore
\[
U_{\mathrm{joint}}=13-4\rho^2.
\]

\paragraph{Marginal audits.}
Write
\[
c=\frac{1+\rho}{2},
\qquad
d=\frac{1-\rho}{2}.
\]
Conditionally on $S_1$,
\[
\mathcal L(Z\mid S_1=0)
=
c\,\delta_0+d\,\delta_{10},
\]
whereas
\[
\mathcal L(Z\mid S_1=1)
=
d\,\delta_4+c\,\delta_6.
\]
The two mixture quantile functions change regime at different probability
levels whenever $\rho\neq0$. If $\rho\geq0$, then $d\leq c$ and
\[
W_2^2\!\left(
\mathcal L(Z\mid S_1=0),
\mathcal L(Z\mid S_1=1)
\right)
=
16d+36(c-d)+16d
=
16+20\rho.
\]
If $\rho\leq0$, the same calculation with the two breakpoints reversed gives
\[
W_2^2\!\left(
\mathcal L(Z\mid S_1=0),
\mathcal L(Z\mid S_1=1)
\right)
=
16-20\rho.
\]
Since $S_1$ is equiprobable,
\[
U_1=4+5|\rho|.
\]

For $S_2$, the conditional laws are
\[
\mathcal L(Z\mid S_2=0)
=
c\,\delta_0+d\,\delta_4,
\]
and
\[
\mathcal L(Z\mid S_2=1)
=
c\,\delta_6+d\,\delta_{10}.
\]
The second distribution is exactly the first shifted by $6$. Hence
\[
W_2^2\!\left(
\mathcal L(Z\mid S_2=0),
\mathcal L(Z\mid S_2=1)
\right)
=
36,
\]
and
\[
U_2=9.
\]
Consequently,
\[
U_{\mathrm{marg}}
=
U_1+U_2
=
13+5|\rho|.
\]
Thus the equality $U_{\mathrm{marg}}=U_{\mathrm{joint}}$ observed under independence is immediately lost when $\rho\neq0$:
\[
U_{\mathrm{marg}}-U_{\mathrm{joint}}
=
5|\rho|+4\rho^2.
\]

\paragraph{Sequential audits.}
Consider first the order $S_1\to S_2$. Conditional on $S_1=0$, revealing
$S_2$ distinguishes the point masses $\delta_0$ and $\delta_{10}$ with
conditional weights $c$ and $d$. Their weighted Wasserstein variance is
\[
100cd.
\]
Conditional on $S_1=1$, the corresponding point masses are $\delta_4$ and
$\delta_6$, giving
\[
4cd.
\]
Averaging over the two values of $S_1$ yields
\[
U_{2\mid1}
=
\frac12(100cd)+\frac12(4cd)
=
52cd.
\]
Since
\[
cd=\frac{1-\rho^2}{4},
\]
we obtain
\[
U_{2\mid1}=13(1-\rho^2).
\]

For the reverse order, the two point masses within each $S_2$ stratum are
separated by $4$. Hence
\[
U_{1\mid2}=4(1-\rho^2).
\]

The two sequential totals are therefore
\[
U_1+U_{2\mid1}
=
17+5|\rho|-13\rho^2,
\]
and
\[
U_2+U_{1\mid2}
=
13-4\rho^2
=
U_{\mathrm{joint}}.
\]
The second equality is specific to this example; it is not a general
decomposition identity for sequential audits. In contrast, for the order
$1\to2$,
\[
U_1+U_{2\mid1}-U_{\mathrm{joint}}
=
4+5|\rho|-9\rho^2
=
(1-|\rho|)(4+9|\rho|).
\]
Thus the sequential total is strictly larger than $U_{\mathrm{joint}}$ for
$|\rho|<1$, except in the reverse order, where equality happens because of
the particular geometry of the four cell values.

\paragraph{The barycentric game.}
Because each full-cell law is degenerate,
\[
Q_S(u)=Z
\qquad\text{for all }u,
\]
and hence
\[
V(A)
=
\operatorname{Var}\!\left(
\mathbb E[Z\mid S_A]
\right).
\]
Clearly,
\[
V(\varnothing)=0.
\]

For $S_1$,
\[
\mathbb E[Z\mid S_1=0]
=
5-5\rho,
\qquad
\mathbb E[Z\mid S_1=1]
=
5+\rho.
\]
Their difference is $6\rho$, and therefore
\[
V(\{1\})=9\rho^2.
\]

For $S_2$,
\[
\mathbb E[Z\mid S_2=0]
=
2-2\rho,
\qquad
\mathbb E[Z\mid S_2=1]
=
8-2\rho.
\]
Their difference is always $6$, so
\[
V(\{2\})=9.
\]

Finally,
\[
V(\{1,2\})
=
U_{\mathrm{joint}}
=
13-4\rho^2.
\]
The complete game is thus
\[
V(\varnothing)=0,\qquad
V(\{1\})=9\rho^2,\qquad
V(\{2\})=9,\qquad
V(\{1,2\})=13-4\rho^2.
\]

This makes the distinction between the actual marginal audit and the
barycentric projection particularly transparent:
\[
U_1=4+5|\rho|,
\qquad
V(\{1\})=9\rho^2.
\]
They agree only under perfect dependence, $|\rho|=1$, whereas under
independence they reduce to the discrepancy discussed in
Appendix~\ref{app:toy-four-cell},
\[
U_1=4,
\qquad
V(\{1\})=0.
\]

\paragraph{Path increments.}
The barycentric increments along $1\to2$ are
\[
H_1
=
V(\{1\})-V(\varnothing)
=
9\rho^2,
\]
and
\[
H_{2\mid1}
=
V(\{1,2\})-V(\{1\})
=
13(1-\rho^2).
\]
Along the reverse order,
\[
H_2=9,
\]
and
\[
H_{1\mid2}
=
V(\{1,2\})-V(\{2\})
=
4(1-\rho^2).
\]
Therefore
\[
H_1+H_{2\mid1}
=
H_2+H_{1\mid2}
=
U_{\mathrm{joint}}.
\]

In this particular deterministic example,
\[
U_{2\mid1}=H_{2\mid1},
\qquad
U_{1\mid2}=H_{1\mid2},
\qquad
U_2=H_2,
\]
while the discrepancy between the two constructions is concentrated in the
first step of the order $1\to2$:
\[
U_1=4+5|\rho|
\qquad\text{versus}\qquad
H_1=9\rho^2.
\]

\paragraph{Harsanyi interaction and redundancy.}
The second-order Harsanyi dividend is
\[
D_{\{1,2\}}
=
V(\{1,2\})-V(\{1\})-V(\{2\}),
\]
hence
\[
D_{\{1,2\}}=4-13\rho^2.
\]
Under independence, $D_{\{1,2\}}=4$, as in the previous example. As the
dependence between the sensitive attributes increases, this interaction
shrinks and becomes negative once
\[
|\rho|>\frac{2}{\sqrt{13}}\approx0.555.
\]
A negative dividend here indicates overlap or redundancy in the information
carried by the two sensitive attributes within the barycentric game.

At perfect dependence,
\[
|\rho|=1,
\]
we obtain
\[
V(\{1\})=V(\{2\})=V(\{1,2\})=9,
\]
and therefore
\[
D_{\{1,2\}}=-9.
\]
Each sensitive attribute then contains, on the observed support, the same
information as the pair.

\paragraph{Shapley allocation.}
For two attributes,
\[
\phi_1
=
\frac12V(\{1\})
+
\frac12\bigl[
V(\{1,2\})-V(\{2\})
\bigr],
\]
which gives
\[
\phi_1
=
2+\frac52\rho^2.
\]
Similarly,
\[
\phi_2
=
11-\frac{13}{2}\rho^2.
\]
Their sum is
\[
\phi_1+\phi_2
=
13-4\rho^2
=
U_{\mathrm{joint}}.
\]

The corresponding shares of joint distributional disparity are
\[
\frac{\phi_1}{U_{\mathrm{joint}}}
=
\frac{2+\frac52\rho^2}{13-4\rho^2},
\qquad
\frac{\phi_2}{U_{\mathrm{joint}}}
=
\frac{11-\frac{13}{2}\rho^2}{13-4\rho^2}.
\]
At independence, $\rho=0$, these reduce to
\[
\frac{\phi_1}{U_{\mathrm{joint}}}
=
\frac{2}{13}
\approx15.4\%,
\qquad
\frac{\phi_2}{U_{\mathrm{joint}}}
=
\frac{11}{13}
\approx84.6\%.
\]
As $|\rho|$ approaches one, the two attributes become informationally
redundant and the allocation converges to
\[
\phi_1=\phi_2=\frac92,
\]
so that each receives exactly $50\%$ of the remaining joint disparity.

\begin{table}[ht]
\centering
\caption{Four-cell example with dependent sensitive attributes. The second
column gives the expressions for
$\rho=\operatorname{Corr}(S_1,S_2)\in[-1,1]$, while the last column recovers
the independent case $\rho=0$.}
\label{tab:toy-dependent}
\begin{tabular}{lcc}
\toprule
Quantity & General $\rho$ & Independence ($\rho=0$)\\
\midrule
$U_{\mathrm{joint}}$
    & $13-4\rho^2$
    & $13$\\
$U_1$
    & $4+5|\rho|$
    & $4$\\
$U_2$
    & $9$
    & $9$\\
$U_{\mathrm{marg}}=U_1+U_2$
    & $13+5|\rho|$
    & $13$\\
$U_{2\mid1}$
    & $13(1-\rho^2)$
    & $13$\\
$U_{1\mid2}$
    & $4(1-\rho^2)$
    & $4$\\
$U_1+U_{2\mid1}$
    & $17+5|\rho|-13\rho^2$
    & $17$\\
$U_2+U_{1\mid2}$
    & $13-4\rho^2$
    & $13$\\
\midrule
$V(\varnothing)$
    & $0$
    & $0$\\
$V(\{1\})$
    & $9\rho^2$
    & $0$\\
$V(\{2\})$
    & $9$
    & $9$\\
$V(\{1,2\})$
    & $13-4\rho^2$
    & $13$\\
\midrule
$H_1$
    & $9\rho^2$
    & $0$\\
$H_{2\mid1}$
    & $13(1-\rho^2)$
    & $13$\\
$H_2$
    & $9$
    & $9$\\
$H_{1\mid2}$
    & $4(1-\rho^2)$
    & $4$\\
\midrule
$D_{\{1,2\}}$
    & $4-13\rho^2$
    & $4$\\
$\phi_1$
    & $2+\dfrac52\rho^2$
    & $2$\\
$\phi_2$
    & $11-\dfrac{13}{2}\rho^2$
    & $11$\\
\bottomrule
\end{tabular}
\end{table}

\begin{table}[ht]
\centering
\caption{Summary of the four-cell example. Sequential $U$-audits use actual
conditional mixture laws, whereas $V$ and its increments are based on
barycentric projections of the full-cell quantile process.}
\label{tab:toy-four-cell}
\begin{tabular}{lc}
\toprule
Quantity & Value\\
\midrule
$U_{\mathrm{joint}}$ & $13$\\
$U_1$ & $4$\\
$U_2$ & $9$\\
$U_{2\mid1}$ & $13$\\
$U_{1\mid2}$ & $4$\\
$U_1+U_{2\mid1}$ & $17$\\
$U_2+U_{1\mid2}$ & $13$\\
\midrule
$V(\varnothing)$ & $0$\\
$V(\{1\})$ & $0$\\
$V(\{2\})$ & $9$\\
$V(\{1,2\})$ & $13$\\
$H_{2\mid1}$ & $13$\\
$H_{1\mid2}$ & $4$\\
$D_{\{1,2\}}$ & $4$\\
$\phi_1$ & $2$\\
$\phi_2$ & $11$\\
\bottomrule
\end{tabular}
\end{table}

\FloatBarrier

\section{Why $W_2$ differs from a linear CDF geometry}
\label{app:w2-vs-cdf}
The mixture--barycenter gap is a consequence of the quantile geometry of $W_2$, not a generic feature of every distributional embedding. To see this, let
\[
\mathcal F_{\bS}(t)=F_{Z\mid\bS}(t)
\]
and, whenever the relevant integral is finite, equip CDFs with the $L^2(dt)$ distance. For every coalition $A$,
\begin{equation}\label{eq:cdf-linear}
\E\{\mathcal F_{\bS}(t)\mid S_A\}=F_{Z\mid S_A}(t)
\qquad\text{for a.e. }t,
\end{equation}
because CDFs are linear under mixture. Hence the corresponding projection game is built from the actual conditional mixture laws themselves. Along a nested reveal path, its projection increments obey an exact Pythagorean chain rule. By contrast, quantiles satisfy
$Q_{\Law(Z\mid S_A)}\neq\E(\mathcal Q\mid S_A)$ in general, which is precisely the obstruction studied in the main text.

\paragraph{A null-player counterexample for the $W_2$ game.}
Let $S_1,S_2$ be independent Bernoulli$(1/2)$ variables and define the four full-cell laws
\[
\nu_{00}=\delta_0,
\qquad \nu_{01}=\delta_2,
\qquad
\nu_{10}=\nu_{11}=\tfrac12\delta_0+\tfrac12\delta_2.
\]
Mixing over $S_2$ gives the same law for $S_1=0$ and $S_1=1$, namely
$\tfrac12\delta_0+\tfrac12\delta_2$; therefore $Z\indep S_1$ and $\mathsf U_1=0$. The full-cell quantiles are
\[
Q_{00}=0,
\qquad Q_{01}=2,
\qquad
Q_{10}=Q_{11}=
\begin{cases}
0,&u<1/2,\\
2,&u\ge1/2.
\end{cases}
\]
The global barycentric quantile is $\bar Q(u)=1/2$ for $u<1/2$ and $3/2$ for $u\ge1/2$. Conditioning the quantile process on $S_1$ gives
\[
M_{\{1\}}(0;u)=1,
\qquad
M_{\{1\}}(1;u)=Q_{10}(u),
\]
so
\[
\mathcal V(\{1\})=\frac14>0.
\]
A direct calculation also gives $\Ujoint=3/4$ and, by terminal equality,
$\mathsf U_{2\mid1}=H_{2\mid1}=\Ujoint-\mathcal V(\{1\})=1/2$. Consequently
\[
\Useq^{(1,2)}=\frac12<\frac34=\Ujoint,
\qquad
\Delta_1^{(1,2)}=-\frac14.
\]
Thus the sequential discrepancy is genuinely signed, and marginal independence of $Z$ from a coordinate does not make that coordinate a null player of the quantile-projection game. This is why the Shapley allocation is interpreted as descriptive information allocation rather than as an independence or causal attribution statement.

\section{Additional theory and proofs}
\label{app:theory}

This appendix develops the geometry behind the main text. The central object is the random full-cell quantile function $\mathcal Q=Q_{\bS}\in\cH=L^2(0,1)$. The logic is worth keeping in mind: the \emph{audit} compares conditional laws and supplies the correct zero set; the \emph{game} projects $\mathcal Q$ onto successively richer sensitive information; and Shapley allocation averages the resulting nonnegative projection increments over reveal orders. The audit and projection constructions answer different questions away from the joint null. Their discrepancy is localized by the results below: they necessarily coincide at the terminal step of every reveal order, but may differ at earlier steps when sensitive coordinates remain to be marginalized.

\subsection{One-dimensional barycenters as Hilbert-space means}
The identity in~\eqref{eq:ujoint} is the geometric hinge of the paper. We record it explicitly because it turns a distributional problem into an ordinary variance decomposition in a function space.

\begin{lemma}[Quantile barycenter identity]\label{lem:app-barycenter}
Let $\nu_1,\ldots,\nu_m\in\mathcal P_2(\mathbb R)$ have quantile functions $Q_1,\ldots,Q_m$ and weights $p_s>0$, $\sum_sp_s=1$. Then
\[
\inf_{\nu\in\mathcal P_2(\mathbb R)}\sum_{s=1}^mp_sW_2^2(\nu_s,\nu)
=
\sum_{s=1}^mp_s\|Q_s-\bar Q\|_{\cH}^2,
\qquad
\bar Q=\sum_{s=1}^mp_sQ_s.
\]
The minimizer is the one-dimensional Wasserstein barycenter with quantile function $\bar Q$.
\end{lemma}

\begin{proof}
For any candidate law $\nu$ with quantile $Q$, one-dimensional optimal transport gives
\[
\sum_sp_sW_2^2(\nu_s,\nu)=\sum_sp_s\|Q_s-Q\|_{\cH}^2.
\]
The Hilbert-space variance identity yields
\[
\sum_sp_s\|Q_s-Q\|_{\cH}^2
=
\sum_sp_s\|Q_s-\bar Q\|_{\cH}^2+\|Q-\bar Q\|_{\cH}^2.
\]
Because a positive weighted average of nondecreasing quantile functions is itself nondecreasing and square-integrable, $\bar Q$ is a valid quantile function. The second term is minimized uniquely up to equality almost everywhere by $Q=\bar Q$, proving the claim.
\end{proof}

\begin{remark}[An $R^2$ analogy in quantile space]\label{rem:app-r2}
The game $\mathcal V(A)=\E\|\E(Q_{\bS}\mid S_A)-\E Q_{\bS}\|_{\cH}^2$ is the Hilbert-valued analogue of explained variance. Revealing $S_A$ predicts the entire conditional quantile curve rather than a scalar response. Thus $\mathcal V(A)/\Ujoint$ can be read as a distributional analogue of an $R^2$ whenever $\Ujoint>0$. The analogy is geometric, not causal: coalitions are information sets, not interventions.
\end{remark}

\subsection{Complete proof of the zero-set statements}
We first prove Proposition~\ref{prop:zeros}. The argument uses only nonnegativity and separation of the underlying distributional discrepancy; Wasserstein geometry becomes essential only for the additive decomposition developed later.

\begin{proof}[Proof of Proposition~\ref{prop:zeros}]
For a categorical variable $T$ with positive-probability cells $t$, its Wasserstein barycentric disparity is a nonnegative weighted sum of terms $W_2^2\{\Law(Z\mid T=t),\nu^\star\}$ at the minimizing reference law $\nu^\star$. This quantity is zero if and only if every positive-probability conditional law equals the same reference law. Averaging those equal laws over $T$ shows that the common law is $\Law(Z)$. Hence the disparity is zero if and only if $Z\indep T$.

Applying this observation to each singleton $S_j$ proves
\[
\Umarg=0\quad\Longleftrightarrow\quad Z\indep S_j\ \text{for every }j.
\]
Applying it to the full profile $T=\bS$ proves
\[
\Ujoint=0\quad\Longleftrightarrow\quad Z\indep\bS.
\]

Now fix an ordering $\pi$. Since all terms in $\Useq^\pi$ are nonnegative, $\Useq^\pi=0$ if and only if every conditional term is zero. For step $k$, zero conditional disparity means that for every positive-probability value $b$ of $S_{A_{k-1}^\pi}$,
\[
\Law\{Z\mid S_{A_{k-1}^\pi}=b,S_{\pi(k)}=a\}
=
\Law\{Z\mid S_{A_{k-1}^\pi}=b\}
\]
for all compatible $a$. Equivalently,
\[
\Law(Z\mid S_{A_k^\pi})=\Law(Z\mid S_{A_{k-1}^\pi})
\quad\text{a.s.}
\]
Iterating from $A_0^\pi=\varnothing$ to $A_d^\pi=[d]$ gives $\Law(Z\mid\bS)=\Law(Z)$, hence $Z\indep\bS$. Conversely, joint independence makes every conditional law in every sequential term equal to $\Law(Z)$, so every term vanishes. This proves the sequential equivalence.
\end{proof}

\subsection{Worst-cell concentration}
\label{app:worst-bound}
A companion statistic is the barycentric worst-cell distance
\[
\Uworst(Z)=\max_{\bs}W_2^2(\nu_{\bs},\bar\nu),
\]
where $\bar\nu$ is the joint Wasserstein barycenter. It has the same joint-demographic-parity zero set as $\Ujoint$ and, unlike a separate minimax-radius statistic, identifies the largest cellwise contribution relative to the same reference used by the average audit.

\begin{proposition}[Average versus barycentric worst cell]\label{prop:app-worst}
Let $p_{\min}=\min_{\bs}p_{\bs}>0$. Then
\[
p_{\min}\Uworst(Z)\le \Ujoint(Z)\le \Uworst(Z).
\]
Hence $\Uworst/\Ujoint\le p_{\min}^{-1}$ whenever $\Ujoint>0$.
\end{proposition}

\begin{proof}
Set $d_{\bs}=W_2^2(\nu_{\bs},\bar\nu)$. Since $\Ujoint=\sum_{\bs}p_{\bs}d_{\bs}$ and $\Uworst=\max_{\bs}d_{\bs}$,
\[
p_{\min}\Uworst\le \sum_{\bs}p_{\bs}d_{\bs}\le \Uworst.
\]
This proves the result.
\end{proof}

On the fixed grid, the plug-in version is
\[
\widehat\Uworst_K=\max_{\bs}\|\widehat q_{\bs}-\widehat{\bar q}\|_K^2,
\qquad
\widehat{\bar q}=\sum_{\bs}\widehat p_{\bs}\widehat q_{\bs},
\]
with empirical weights renormalized if the audit is defined on a retained support. No Chebyshev-center optimization is performed. Consistency follows from the same cellwise quantile and probability consistency used for $\widehat\Ujoint_K$.

\begin{corollary}[Collapse of all decomposable quantities at the joint null]\label{cor:app-null-collapse}
The following are equivalent:
\[
Z\indep\bS,
\qquad
\Ujoint=0,
\qquad
\mathcal V([d])=0,
\qquad
M_A=M_\varnothing\ \text{a.s. for every }A\subseteq[d].
\]
Under these conditions every path increment and every Shapley allocation is zero. Conversely, if all Shapley allocations vanish, then $Z\indep\bS$.
\end{corollary}

\begin{proof}
By Lemma~\ref{lem:app-barycenter}, $\Ujoint=\E\|Q_{\bS}-\E Q_{\bS}\|^2$, so $\Ujoint=0$ exactly when $Q_{\bS}$ is almost surely constant, equivalently when all positive-probability full-cell laws coincide. This is joint demographic parity. If $Q_{\bS}$ is constant, every conditional projection $M_A$ equals $M_\varnothing$; the converse follows by taking $A=[d]$. The statements for increments and Shapley values follow from Theorem~\ref{thm:main}. Conversely, nonnegative Shapley values summing to $\Ujoint$ can all vanish only when $\Ujoint=0$.
\end{proof}

The zero-set conclusions extend verbatim to any nonnegative separating discrepancy $D(\mu,\nu)$: replacing $W_2^2$ by $D$ preserves the marginal, joint, worst-cell, and complete sequential certification statements. What is special to one-dimensional $W_2^2$ is the additional Hilbert geometry that makes a fixed positive joint disparity exactly decomposable.

\subsection{Why the conditional audit is not a chain rule}
For $(z_{00},z_{01},z_{10},z_{11})=(0,10,4,6)$, the full-cell barycenter is $5$ and
\[
\Ujoint=\tfrac14\{(0-5)^2+(10-5)^2+(4-5)^2+(6-5)^2\}=13.
\]
For $S_1=0$, the full-cell values are $0,10$ and their within-stratum barycentric dispersion is $25$; for $S_1=1$, the values are $4,6$ and the corresponding dispersion is $1$. Hence the second conditional step contributes $(25+1)/2=13$. The two coarsened laws given $S_1$ are
\[
\tfrac12(\delta_0+\delta_{10}),\qquad \tfrac12(\delta_4+\delta_6),
\]
whose equal-weight Wasserstein disparity is $4$. Thus $\Useq^{(1,2)}=4+13=17\ne13$.

The calculation isolates the obstruction. A conditional audit first \emph{mixes laws} inside the coarser strata and then compares those mixtures. The decomposition below first embeds every full-cell law by its quantile curve and then \emph{averages those curves}. Since the quantile of a mixture is generally not the mixture of the quantiles, the two operations need not agree. At zero they certify the same joint independence statement; away from zero they measure different positive quantities.

\subsection{Proofs for the sequential discrepancy results}
\label{app:sequential_discrepancy}

\begin{proof}[Proof of Proposition~\ref{prop:terminal-equivalence}]
Fix $j\in[d]$ and let
\[
    A=[d]\setminus\{j\}.
\]
Recall that
\[
    Q_{\bS}
    =
    Q_{\Law(Z\mid \bS)}
    \in \cH=L^2(0,1),
\]
and
\[
    M_A
    =
    \E[Q_{\bS}\mid S_A].
\]

Fix a stratum $S_A=a$. Since $A$ contains every sensitive coordinate except $j$, conditioning further on $S_j=s_j$ specifies a full sensitive cell. Hence
\[
    \Law(Z\mid S_A=a,S_j=s_j)
    =
    \nu_{a,s_j},
\]
with quantile function $Q_{a,s_j}$.

The conditional Wasserstein audit within stratum $a$ is therefore
\[
    \inf_{\nu\in\mathcal P_2(\mathbb R)}
    \sum_{s_j}
    \Pr(S_j=s_j\mid S_A=a)
    W_2^2(\nu_{a,s_j},\nu).
\]
In one dimension, the Wasserstein barycenter has quantile function
\[
    \bar Q_a
    =
    \sum_{s_j}
    \Pr(S_j=s_j\mid S_A=a)\,
    Q_{a,s_j}.
\]
But, by definition of conditional expectation,
\[
    \bar Q_a
    =
    \E[Q_{\bS}\mid S_A=a]
    =
    M_A(a).
\]
Consequently, the within-stratum conditional disparity is
\[
    \sum_{s_j}
    \Pr(S_j=s_j\mid S_A=a)
    \left\|
        Q_{a,s_j}-M_A(a)
    \right\|_H^2.
\]

Averaging over $S_A$ gives
\[
    \mathsf U_{\{j\}\mid A}
    =
    \E\left[
        \|Q_{\bS}-M_A\|_{\cH}^2
    \right].
\]

Since $M_A=\E[Q_S\mid S_A]$ is the orthogonal projection of $Q_S$ onto the closed subspace of $S_A$-measurable $H$-valued random variables, the Hilbert-space Pythagorean identity yields
\[
    \E\|Q_{\bS}-M_\varnothing\|_{\cH}^2
    =
    \E\|M_A-M_\varnothing\|_{\cH}^2
    +
    \E\|Q_{\bS}-M_A\|_{\cH}^2.
\]
Using
\[
    \mathcal V([d])
    =
    \E\|Q_{\bS}-M_\varnothing\|_{\cH}^2
    =
    \Ujoint
\]
and
\[
    \mathcal V(A)
    =
    \E\|M_A-M_\varnothing\|_{\cH}^2,
\]
we obtain
\[
    \mathsf U_{\{j\}\mid A}
    =
    \mathcal V([d])-V(A).
\]
Finally, by definition,
\[
    H_{j\mid A}
    =
    \mathcal V(A\cup\{j\})-V(A)
    =
    \mathcal V([d])-V(A),
\]
which proves
\[
    \mathsf U_{\{j\}\mid A}
    =
    H_{j\mid A}
    =
    \mathcal V([d])-V(A).
\]
No independence assumption is used.
\end{proof}

\begin{proof}[Proof of Proposition~\ref{prop:sequential-discrepancy}]
By definition, for an ordering $\pi$ over the sensitive attributes
\[
    \Useq^\pi
    =
    \sum_{k=1}^d
    \mathsf U_{\{\pi(k)\}\mid A_{k-1}^\pi}.
\]
On the other hand, the barycentric increments telescope:
\begin{align*}
    \Ujoint
    &=
    \mathcal V([d])-V(\varnothing)\\
    &=
    \sum_{k=1}^d
    \left\{
        V(A_k^\pi)-V(A_{k-1}^\pi)
    \right\}\\
    &=
    \sum_{k=1}^d
    H_{\pi(k)\mid A_{k-1}^\pi}.
\end{align*}
Subtracting the two identities gives
\[
    \Useq^\pi-U_{\mathrm{joint}}
    =
    \sum_{k=1}^d
    \left[
        \mathsf U_{\{\pi(k)\}\mid A_{k-1}^\pi}
        -
        H_{\pi(k)\mid A_{k-1}^\pi}
    \right]
    =
    \sum_{k=1}^d\Delta_k^\pi.
\]

At the final step,
\[
    A_{d-1}^\pi
    =
    [d]\setminus\{\pi(d)\}.
\]
Proposition~\ref{prop:terminal-equivalence} therefore implies
\[
    \mathsf U_{\{\pi(d)\}\mid A_{d-1}^\pi}
    =
    H_{\pi(d)\mid A_{d-1}^\pi},
\]
and hence
\[
    \Delta_d^\pi=0.
\]
It follows that
\[
    \Useq^\pi-U_{\mathrm{joint}}
    =
    \sum_{k=1}^{d-1}\Delta_k^\pi.
\]
\end{proof}

\begin{proof}[Proof of Corollary~\ref{cor:two-sensitive}]
When $d=2$, Proposition~\ref{prop:terminal-equivalence} applied with
$j=2$ and $A=\{1\}$ gives
\[
    \mathsf U_{2\mid1}
    =
    H_{2\mid1}
    =
    \mathcal V(\{1,2\})-V(\{1\}).
\]
Exchanging the roles of the two attributes gives
\[
    \mathsf U_{1\mid2}
    =
    H_{1\mid2}
    =
    \mathcal V(\{1,2\})-V(\{2\}).
\]

Therefore,
\begin{align*}
    \Useq^{(1,2)}
    &=
    \mathsf U_1+U_{2\mid1}\\
    &=
    \mathsf U_1+V(\{1,2\})-V(\{1\})\\
    &=
    \Ujoint
    +
    \bigl\{U_1-V(\{1\})\bigr\},
\end{align*}
where we used $V(\{1,2\})=U_{\mathrm{joint}}$.
The second identity follows symmetrically:
\[
    \Useq^{(2,1)}
    =
    \Ujoint
    +
    \bigl\{U_2-V(\{2\})\bigr\}.
\]
\end{proof}

\subsection{Full proof and interpretation of Theorem~\ref{thm:main}}
\label{app:proof-main}
Let $\mathcal K=L^2(\Omega;\cH)$ with inner product $\langle X,Y\rangle_{\mathcal K}=\E\langle X,Y\rangle_{\cH}$. For $A\subseteq[d]$, $M_A=\E(Q_{\bS}\mid S_A)$ is the orthogonal projection of $Q_{\bS}$ onto the closed subspace of $S_A$-measurable elements of $\mathcal K$.

\begin{proof}[Proof of Theorem~\ref{thm:main}]
If $A\subseteq B$, the tower property gives
\[
M_A=\E(M_B\mid S_A).
\]
Therefore $M_B-M_A$ is orthogonal to every square-integrable $S_A$-measurable $\cH$-valued random variable, and in particular to $M_A-M_\varnothing$. Since
\[
M_B-M_\varnothing=(M_B-M_A)+(M_A-M_\varnothing),
\]
Pythagoras in $\mathcal K$ yields
\[
\mathcal V(B)=\mathcal V(A)+\E\|M_B-M_A\|_{\cH}^2.
\]
This proves the increment identity in Theorem~\ref{thm:main} and monotonicity. Also $M_\varnothing=\E Q_{\bS}$ and $M_{[d]}=Q_{\bS}$, so Lemma~\ref{lem:app-barycenter} gives
\[
\mathcal V(\varnothing)=0,
\qquad
\mathcal V([d])=\E\|Q_{\bS}-\E Q_{\bS}\|^2=\Ujoint.
\]

For a permutation $\pi$, apply the preceding identity to the nested chain
\[
\varnothing=A_0^\pi\subset A_1^\pi\subset\cdots\subset A_d^\pi=[d].
\]
Then
\[
\mathcal V(A_k^\pi)-\mathcal V(A_{k-1}^\pi)
=
\E\|M_{A_k^\pi}-M_{A_{k-1}^\pi}\|^2,
\]
and summing over $k$ telescopes from $0$ to $\Ujoint$, equivalently yielding~\eqref{eq:H-telescope}.

Finally, the Shapley formula is a convex average of coalition increments $\mathcal V(A\cup\{j\})-\mathcal V(A)$. Monotonicity makes every such increment nonnegative, hence $\phi_j\ge0$. Standard Shapley efficiency, or equivalently averaging the telescoping identity over all $d!$ permutations, gives
\[
\sum_{j=1}^d\phi_j=\mathcal V([d])-\mathcal V(\varnothing)=\Ujoint.
\]
\end{proof}

\begin{corollary}[Edge-wise zero criterion]\label{cor:app-edge-zero}
For $j\notin A$,
\[
\mathcal V(A\cup\{j\})-\mathcal V(A)=0
\quad\Longleftrightarrow\quad
M_{A\cup\{j\}}=M_A\ \text{a.s.}
\]
Thus an edge of the coalition lattice is zero exactly when revealing $S_j$ adds no information about the full-cell quantile curve beyond $S_A$.
\end{corollary}

\begin{proof}
By Theorem~\ref{thm:main}, the edge increment is $\E\|M_{A\cup\{j\}}-M_A\|^2$, which vanishes exactly when the two projections agree almost surely.
\end{proof}

\begin{remark}[A zero edge is weaker than conditional independence]\label{rem:app-zero-edge}
The criterion in Corollary~\ref{cor:app-edge-zero} is a barycentric information statement. In general it does \emph{not} imply $Z\indep S_j\mid S_A$, because $M_A$ averages full-cell quantile functions whereas the conditional law $\Law(Z\mid S_A)$ is a probabilistic mixture of full-cell laws. This is the same mixture-versus-barycenter distinction that prevents a Wasserstein chain rule.
\end{remark}

\subsection{Independent coordinates: full ANOVA proof}
Assume now that $S_1,\ldots,S_d$ are mutually independent. Define recursively
\[
h_\varnothing=M_\varnothing,
\qquad
h_A(S_A)=M_A-\sum_{B\subsetneq A}h_B(S_B),\quad A\ne\varnothing,
\]
and set $\mathsf I_A=\E\|h_A(S_A)\|_{\cH}^2$.

\begin{proposition}[Orthogonal barycentric ANOVA]\label{prop:app-anova}
The components $(h_A)_{A\subseteq[d]}$ satisfy
\[
Q_{\bS}=h_\varnothing+\sum_{\varnothing\ne A\subseteq[d]}h_A(S_A),
\]
are mutually orthogonal in $L^2(\Omega;\cH)$, and for every coalition $C$,
\[
M_C=h_\varnothing+\sum_{\varnothing\ne A\subseteq C}h_A(S_A).
\]
Consequently,
\[
\mathcal V(C)=\sum_{\varnothing\ne A\subseteq C}\mathsf I_A,
\qquad
\Ujoint=\sum_{A\ne\varnothing}\mathsf I_A.
\]
\end{proposition}

\begin{proof}
The recursion immediately gives $M_A=\sum_{B\subseteq A}h_B(S_B)$. Taking $A=[d]$ and using $M_{[d]}=Q_{\bS}$ yields the full expansion.

We next establish the standard degeneracy property. Fix nonempty $A$ and $j\in A$. By induction on $|A|$, using independence of the coordinates and the recursion,
\[
\E\{h_A(S_A)\mid S_{A\setminus\{j\}}\}=0.
\]
Indeed, condition $M_A=\E(Q_{\bS}\mid S_A)$ on $S_{A\setminus\{j\}}$ to obtain $M_{A\setminus\{j\}}$; after substituting the recursive expansions, all terms not containing $j$ cancel and all proper terms containing $j$ have zero conditional expectation by the induction hypothesis.

Now take $A\ne B$ and choose $j\in A\setminus B$ without loss of generality. Because $h_B(S_B)$ is measurable with respect to the coordinates other than $S_j$, independence and the degeneracy property imply
\[
\E\langle h_A(S_A),h_B(S_B)\rangle
=
\E\Bigl\langle \E\{h_A(S_A)\mid S_{[d]\setminus\{j\}}\},h_B(S_B)\Bigr\rangle=0.
\]
Thus the components are mutually orthogonal.

Finally, condition the full expansion on $S_C$. If $A\subseteq C$, $h_A(S_A)$ is measurable and remains unchanged. If $A\not\subseteq C$, choose $j\in A\setminus C$; integrating out $S_j$ and using degeneracy makes that component vanish. Hence
\[
M_C=h_\varnothing+\sum_{\varnothing\ne A\subseteq C}h_A(S_A).
\]
Subtracting $M_\varnothing=h_\varnothing$ and applying orthogonality gives the formula for $\mathcal V(C)$; taking $C=[d]$ gives the decomposition of $\Ujoint$.
\end{proof}

The proposition gives a concrete answer to ``where the joint disparity lives'': first-order energies are main effects, second-order energies are genuinely pairwise components, and so on. Barycentric reveal paths and the Shapley rule differ only in how they allocate these same orthogonal energies; this statement does not apply to ordinary conditional Wasserstein audits.

\begin{proposition}[Sequential allocation of interactions]\label{prop:app-seq-anova}
For an ordering $\pi$ and $A_k^\pi=\{\pi(1),\ldots,\pi(k)\}$,
\[
\mathcal V(A_k^\pi)-\mathcal V(A_{k-1}^\pi)
=
\sum_{\substack{B\subseteq A_k^\pi\\ \pi(k)\in B}}\mathsf I_B.
\]
Hence each interaction energy is assigned entirely to the last member of that interaction to enter the ordering.
\end{proposition}

\begin{proof}
By Proposition~\ref{prop:app-anova}, subtract
\[
\mathcal V(A_{k-1}^\pi)=\sum_{\varnothing\ne B\subseteq A_{k-1}^\pi}\mathsf I_B
\]
from the corresponding sum over $A_k^\pi$. The surviving subsets are precisely those containing the newly added coordinate $\pi(k)$.
\end{proof}

\begin{proof}[Proof of the Shapley sharing identity~\eqref{eq:H-anova}]
Fix an interaction set $A$ containing $j$. By Proposition~\ref{prop:app-seq-anova}, the energy $\mathsf I_A$ enters the marginal contribution of $j$ exactly for permutations in which $j$ is the last member of $A$ to appear. Under a uniformly random permutation, each of the $|A|$ members of $A$ is equally likely to be last, so this occurs with probability $1/|A|$. Averaging the path increments over all permutations therefore gives
\[
\phi_j=\sum_{A\ni j}\frac{\mathsf I_A}{|A|}.
\]
Summing over $j$ allocates every $\mathsf I_A$ exactly once and recovers $\Ujoint$.
\end{proof}

\begin{proposition}[Harsanyi--ANOVA identity]\label{prop:app-harsanyi}
Define the Harsanyi dividend of a nonempty coalition $A$ by
\[
\mathcal I(A)=\sum_{B\subseteq A}(-1)^{|A|-|B|}\mathcal V(B).
\]
Under mutual independence, $\mathcal I(A)=\mathsf I_A$ for every nonempty $A$.
\end{proposition}

\begin{proof}
Use $\mathcal V(B)=\sum_{\varnothing\ne C\subseteq B}\mathsf I_C$ and exchange the finite sums:
\[
\mathcal I(A)
=
\sum_{\varnothing\ne C\subseteq A}\mathsf I_C
\sum_{B:\,C\subseteq B\subseteq A}(-1)^{|A|-|B|}.
\]
The inner sum equals $(1-1)^{|A\setminus C|}$, hence is one if $C=A$ and zero otherwise. Only $\mathsf I_A$ remains.
\end{proof}

A useful companion to Shapley is the total interaction-inclusive effect of coordinate $j$,
\[
\mathsf T_j:=\Ujoint-\mathcal V([d]\setminus\{j\}).
\]
Under independence, Proposition~\ref{prop:app-anova} gives $\mathsf T_j=\sum_{A\ni j}\mathsf I_A$.

\begin{corollary}[Main-effect--Shapley--total-effect bracket]\label{cor:app-shapley-bracket}
Under independence,
\[
\mathcal V(\{j\})=\mathsf I_{\{j\}}
\ \le\ 
\phi_j
\ \le\ 
\mathsf T_j=\Ujoint-\mathcal V([d]\setminus\{j\}).
\]
Thus Shapley lies between the isolated main effect of $S_j$ and the full amount of barycentric variation carried by all interactions involving $S_j$.
\end{corollary}

\begin{proof}
From~\eqref{eq:H-anova},
\[
\phi_j=\mathsf I_{\{j\}}+\sum_{\substack{A\ni j\\|A|\ge2}}\frac{\mathsf I_A}{|A|}.
\]
All interaction energies are nonnegative and $1/|A|\in(0,1]$, which gives both inequalities.
\end{proof}

For $d=3$, the whole geometry can be read from seven nonnegative numbers:
\[
\Ujoint=\mathsf I_1+\mathsf I_2+\mathsf I_3+\mathsf I_{12}+\mathsf I_{13}+\mathsf I_{23}+\mathsf I_{123},
\]
while, for example,
\[
\phi_1=\mathsf I_1+\tfrac12(\mathsf I_{12}+\mathsf I_{13})+\tfrac13\mathsf I_{123}.
\]
Along the path $1\to2\to3$, the same energies are instead grouped as $\mathsf I_1$, then $\mathsf I_2+\mathsf I_{12}$, then $\mathsf I_3+\mathsf I_{13}+\mathsf I_{23}+\mathsf I_{123}$. This is the simplest way to see the distinction between a sequential narrative and an order-free allocation.

\subsection{Lower-order parity is a law-level statement}
For $m\in\{1,\ldots,d\}$, say that $Z$ satisfies $m$-way demographic parity if $Z\indep S_A$ for every nonempty $A\subseteq[d]$ with $|A|\le m$. The hierarchy is strict in general: $d$-way parity implies $(d-1)$-way parity, but the converse can fail through pure higher-order interactions.

For every bounded measurable $\varphi:\mathbb R\to\mathbb R$, set
\[
g_\varphi(\bs)=\E\{\varphi(Z)\mid\bS=\bs\}-\E\varphi(Z).
\]
Under independence, write its scalar Hoeffding decomposition as $g_\varphi(\bS)=\sum_{A\ne\varnothing}g_{\varphi,A}(S_A)$.

\begin{proposition}[Lower-order parity and scalar ANOVA]\label{prop:app-lower-dp}
Under mutual independence, $Z$ satisfies $m$-way demographic parity if and only if, for every bounded measurable $\varphi$,
\[
g_{\varphi,A}=0
\qquad\text{for every }A\text{ with }1\le |A|\le m.
\]
\end{proposition}

\begin{proof}
For any coalition $C$,
\[
\E\{g_\varphi(\bS)\mid S_C\}
=
\E\{\varphi(Z)\mid S_C\}-\E\varphi(Z).
\]
The scalar Hoeffding projection identity gives
\[
\E\{g_\varphi(\bS)\mid S_C\}
=
\sum_{\varnothing\ne A\subseteq C}g_{\varphi,A}(S_A).
\]
If $Z\indep S_C$ for every $|C|\le m$, the left-hand side is zero for every bounded $\varphi$. Induction on $|C|$ then removes the lower-order terms and yields $g_{\varphi,C}=0$ for every $|C|\le m$. Conversely, if all components of order at most $m$ vanish, the displayed conditional expectation is zero for every $|C|\le m$ and every bounded $\varphi$. Equality of all bounded conditional expectations is equivalent to $\Law(Z\mid S_C)=\Law(Z)$, hence to $Z\indep S_C$.
\end{proof}

This proposition and Proposition~\ref{prop:app-anova} have the same interaction-order intuition but are not the same statement. Lower-order demographic parity concerns \emph{mixtures of laws} after unobserved coordinates are integrated out; $\mathcal V(A)$ concerns the \emph{Wasserstein barycenter} of the full-cell quantile curves compatible with $S_A$. Therefore the ANOVA energies decompose the chosen joint barycentric disparity but are not, by themselves, certificates of lower-order demographic parity.

\subsection{Dependent coordinates: Gaussian formulas and suppression}
\label{app:dependent-gaussian}
The game remains monotone under dependence, but its orthogonal ANOVA interpretation disappears. A Gaussian model provides an exact analogue of sequential regression and makes the role of redundancy and suppression explicit.

\begin{proposition}[Gaussian coalition and residual-increment formulas]\label{prop:app-gaussian-general}
Let $\bS\sim N(0,\Sigma)$ with nonsingular relevant principal submatrices and
\[
Z=\beta^\top\bS+\varepsilon,
\qquad
\varepsilon\indep\bS,
\qquad
\varepsilon\sim N(0,\sigma^2).
\]
For $A\subseteq[d]$, set $c_A=\Cov(S_A,Z)=\Sigma_{A,[d]}\beta$. Then
\[
\mathcal V(A)=c_A^\top\Sigma_{AA}^{-1}c_A.
\]
For $j\notin A$, define
\[
s_{j\mid A}=\Var\{S_j-\E(S_j\mid S_A)\},
\qquad
c_{j\mid A}=\Cov\{Z-\E(Z\mid S_A),S_j-\E(S_j\mid S_A)\}.
\]
Then
\[
\mathcal V(A\cup\{j\})-\mathcal V(A)=\frac{c_{j\mid A}^2}{s_{j\mid A}}.
\]
\end{proposition}

\begin{proof}
Joint Gaussianity gives
\[
\E(Z\mid S_A)=c_A^\top\Sigma_{AA}^{-1}S_A
\]
and a conditional variance that does not depend on the realized value of $S_A$. Conditional score laws therefore differ only by location, so their one-dimensional $W_2^2$ barycentric dispersion equals the variance of the conditional mean:
\[
\mathcal V(A)=\Var\{\E(Z\mid S_A)\}=c_A^\top\Sigma_{AA}^{-1}c_A.
\]
For the increment, residualize both $Z$ and $S_j$ on $S_A$:
\[
R_Z^A=Z-\E(Z\mid S_A),
\qquad
R_j^A=S_j-\E(S_j\mid S_A).
\]
In a joint Gaussian vector, the conditional linear predictor of $R_Z^A$ from $R_j^A$ is
\[
\E(R_Z^A\mid R_j^A)=\frac{c_{j\mid A}}{s_{j\mid A}}R_j^A.
\]
Adding $S_j$ to $S_A$ therefore increases explained variance by
\[
\Var\{\E(R_Z^A\mid R_j^A)\}=\frac{c_{j\mid A}^2}{s_{j\mid A}},
\]
which is exactly $\mathcal V(A\cup\{j\})-\mathcal V(A)$ by the nested-projection identity of Theorem~\ref{thm:main}.
\end{proof}

Thus the edge increment in the Gaussian model is a squared partial-correlation numerator divided by residual variance. It is the distributional-Wasserstein analogue of the extra sum of squares from adding one regressor after the current coalition.

\begin{proposition}[Bivariate conditional amplification]\label{prop:app-gaussian-bivar}
Let $(S_1,S_2)$ be centered bivariate Gaussian with unit variances and correlation $|\rho|<1$, and let
\[
Z=\beta_1S_1+\beta_2S_2+\varepsilon,
\qquad \varepsilon\indep(S_1,S_2).
\]
Then
\[
\mathcal V(\{2\})=(\beta_2+\rho\beta_1)^2,
\qquad
H_{2\mid1}=\beta_2^2(1-\rho^2),
\]
with the symmetric formulas obtained by exchanging $1$ and $2$. Moreover
\[
H_{2\mid1}>\mathcal V(\{2\})
\quad\Longleftrightarrow\quad
-2\rho\beta_1\beta_2-\rho^2(\beta_1^2+\beta_2^2)>0.
\]
If $\rho\in(0,1)$ and $\beta_1=-\beta_2\ne0$, then
\[
\frac{H_{2\mid1}}{\mathcal V(\{2\})}=\frac{1+\rho}{1-\rho}.
\]
\end{proposition}

\begin{proof}
Because $\E(S_1\mid S_2)=\rho S_2$,
\[
\E(Z\mid S_2)=(\beta_2+\rho\beta_1)S_2,
\]
so Proposition~\ref{prop:app-gaussian-general} gives $\mathcal V(\{2\})=(\beta_2+\rho\beta_1)^2$. After conditioning on $S_1$, the residual part of $S_2$ has variance $1-\rho^2$, and the coefficient of that residual in $Z$ is $\beta_2$, giving $H_{2\mid1}=\beta_2^2(1-\rho^2)$. Subtracting the two expressions yields the stated inequality condition. When $\beta_1=-\beta_2$, cancellation gives $\mathcal V(\{2\})=\beta_2^2(1-\rho)^2$ and $H_{2\mid1}=\beta_2^2(1-\rho)(1+\rho)$, whose ratio is $(1+\rho)/(1-\rho)$.
\end{proof}

\begin{figure}[h]
    \centering
    \includegraphics[width=0.62\linewidth]{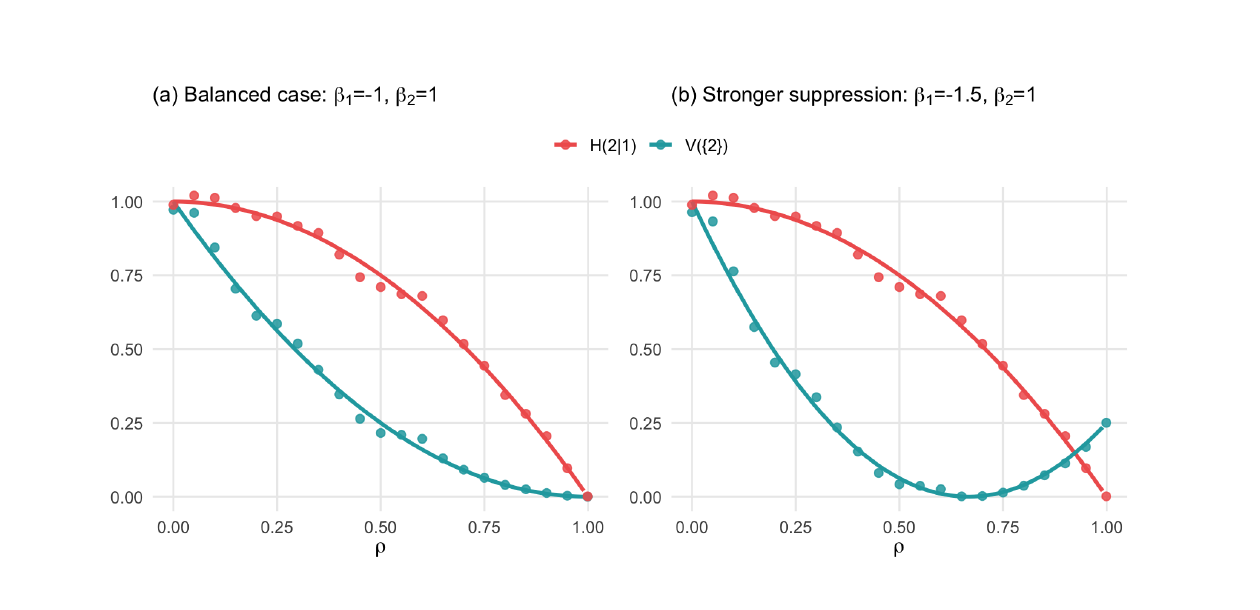}
    \caption{Gaussian location benchmark. Population coalition values and conditional projection increments illustrate marginal cancellation under correlated sensitive coordinates.}
    \label{fig:gauss}
\end{figure}

\begin{remark}[Repair can reveal another marginal disparity]\label{rem:app-residual-repair}
In the same Gaussian model, residualize against $S_1$:
\[
\widetilde Z_1=Z-\E(Z\mid S_1)=\beta_2(S_2-\rho S_1)+\varepsilon.
\]
Then $\widetilde Z_1\indep S_1$, but
\[
\mathcal V_{\widetilde Z_1}(\{2\})=\beta_2^2(1-\rho^2)^2.
\]
This can exceed the original $\mathcal V_Z(\{2\})=(\beta_2+\rho\beta_1)^2$; for $\beta_1=-\beta_2/\rho$, the original marginal value is exactly zero while the residualized one is positive. The same suppression geometry therefore appears not only in auditing but also when one marginal dependence is removed.
\end{remark}

Under dependence, these formulas should remain descriptive. A coalition $A$ means ``reveal $S_A$'' rather than ``intervene on $S_A$''. Monotonicity and Shapley efficiency survive because they are projection properties; orthogonal interaction energies do not.

\section{Statistical supplement}
\label{app:statistics}
\label{app:stat}

\subsection{Proofs of the main-text statistical propositions}
\begin{proof}[Proof of Proposition~\ref{prop:consistency}]
For each positive-probability full cell $\bs$, the law of large numbers gives $\widehat p_{\bs}\to p_{\bs}$ in probability and $n_{\bs}\to\infty$ in probability. Continuity and strict increase of the cell CDF at the selected grid points imply empirical-quantile consistency,
\[
\widehat Q_{\bs}(u_k)\to Q_{\bs}(u_k)
\qquad(k=1,\ldots,K).
\]
There are finitely many cells and grid points, so this convergence holds jointly. Every empirical coalition weight $\widehat p(\bs\mid S_A=a)$ is a continuous ratio of cell probabilities on the interior where $p_A(a)>0$, hence $\widehat M_A(a;u_k)\to M_A(a;u_k)$ jointly. Equation~\eqref{eq:estimator} is a finite polynomial/rational combination of these quantities, so the continuous mapping theorem yields
\[
\widehat{\mathcal V}_K(A)\to\mathcal V_K(A)
\]
for every $A$. The Shapley estimator is a finite linear combination of coalition values, which proves its consistency as well. The regular $\sqrt n$ statement follows by combining Proposition~\ref{prop:app-clt} with the delta method in Proposition~\ref{prop:app-delta}; the exact-null statement is Proposition~\ref{prop:app-null}.
\end{proof}

\begin{proof}[Proof of Proposition~\ref{prop:perm}]
Under $H_0:Z\indep\bS$, the joint sample density or mass function factors as
\[
\prod_{i=1}^n p_Z(z_i)p_S(s_i).
\]
Conditional on the multiset of observed scores $(z_1,\ldots,z_n)$ and on the observed sensitive profiles $(s_1,\ldots,s_n)$, every reassignment of the scores to the profiles therefore has the same probability. The observed pairing is exchangeable over the permutation group. Consequently, for any statistic recomputed after each reassignment, the conditional distribution over all $n!$ permutations is its exact null randomization distribution. Ranking the observed statistic among this distribution gives a finite-sample valid randomization test (with the usual discreteness at the boundary). With $B$ sampled permutations, the standard corrected Monte Carlo p-value
\[
\widehat p=\frac{1+\sum_{b=1}^B\mathbf1\{T^{(b)}\ge T^{\rm obs}\}}{B+1}
\]
is super-uniform and therefore retains valid finite-resolution randomization calibration.
\end{proof}

\subsection{Finite-grid target and grid approximation}
\label{app:grid-approx}
For a fixed grid $u_1,\ldots,u_K$, write $\|f\|_K^2=K^{-1}\sum_{k=1}^Kf(u_k)^2$ and let $\mathcal V_K(A)$ be the corresponding finite-grid coalition value. The consistency proposition in the main text targets $\mathcal V_K(A)$; approximation to the integrated quantity is a separate numerical issue.

\begin{proposition}[Grid approximation]\label{prop:app-grid}
For $a\in\mathcal S_A^+$ define $g_{A,a}(u)=\{M_A(a;u)-M_\varnothing(u)\}^2$. If this finite family is uniformly bounded and uniformly H\"older with exponent $\alpha\in(0,1]$, then on an equally spaced grid
\[
|\mathcal V_K(A)-\mathcal V(A)|\le C_AK^{-\alpha}.
\]
Hence any fixed linear combination of coalition values, including Shapley and martingale quantities, has grid error $O(K^{-\alpha})$.
\end{proposition}
\begin{proof}
For each retained $a$, the difference between the integral of $g_{A,a}$ and its equally spaced Riemann sum is bounded by a constant times the mesh width raised to $\alpha$, hence by $C_{A,a}K^{-\alpha}$. The number of coalition cells is finite and the weights $p_A(a)$ sum to one, so summing the cellwise bounds gives $|\mathcal V_K(A)-\mathcal V(A)|\le C_AK^{-\alpha}$. Any martingale increment or Shapley value is a fixed finite linear combination of coalition values, so the same rate follows by the triangle inequality.
\end{proof}

\subsection{Cellwise quantile CLT} \label{app:subsec-clt}
These fixed-grid limits follow from standard empirical quantile-process asymptotics; Wasserstein statistics based on such quantile-process approximations have a long history, including \citet{delbarrio1999goodness}.
Let the positive-probability full cells be indexed by $s=1,\ldots,m$, and write $q_s=(Q_s(u_1),\ldots,Q_s(u_K))$. Suppose each cell CDF has a density $f_s$ that is continuous and strictly positive at the grid quantiles. 

\begin{proposition}[Cellwise quantile CLT]\label{prop:app-clt}
The vector of empirical cell proportions and empirical cell quantiles satisfies a joint finite-dimensional CLT. In particular,
\[
\sqrt n(\widehat q_s-q_s)\Rightarrow G_s^q,
\]
with covariance
\[
\Cov(G^q_{s,\ell},G^q_{s,\ell'})=
\frac{\min(u_\ell,u_{\ell'})-u_\ell u_{\ell'}}
{p_s f_s(Q_s(u_\ell))f_s(Q_s(u_{\ell'}))}.
\]
\end{proposition}
\begin{proof}
For a fixed cell $s$ and grid level $u_\ell$, the conditional Bahadur representation can be written on the $\sqrt n$ scale as
\[
\sqrt n\{\widehat Q_s(u_\ell)-Q_s(u_\ell)\}
=
\frac{1}{p_sf_s(Q_s(u_\ell))}\frac1{\sqrt n}\sum_{i=1}^n
\mathbf1\{S_i=s\}\bigl[u_\ell-\mathbf1\{Z_i\le Q_s(u_\ell)\}\bigr]+o_{\Pbb}(1).
\]
Together with the usual multinomial influence functions for the cell proportions, this gives a finite collection of mean-zero i.i.d. influence functions. The multivariate central limit theorem therefore yields joint Gaussian convergence. For two grid levels in the same cell, the covariance of the bracketed Bernoulli terms conditional on $S=s$ is $\min(u_\ell,u_{\ell'})-u_\ell u_{\ell'}$; multiplying by $p_s$ from the cell indicator and dividing by $p_s^2f_sf_s$ gives the displayed covariance. Influence functions from distinct cells have zero cross-covariance because their cell indicators are disjoint and each bracketed term has conditional mean zero.
\end{proof}
The factor $1/p_s$ exposes the rare-cell cost directly: even with a fixed total sample size, conditional quantiles in small intersectional cells have high first-order variance.

\subsection{Delta method away from degeneracy}
Let $\theta=((p_s,q_s))_{s=1}^m$ and let $T_K(\theta)$ be any finite vector obtained by smooth finite combinations of coalition values: joint disparity, martingale increments, Shapley values, or Harsanyi contrasts.

\begin{proposition}[Delta method]\label{prop:app-delta}
If $T_K$ is differentiable at $\theta$ with Jacobian $J_K(\theta)$ and the conditions of Proposition~\ref{prop:app-clt} hold, then
\[
\sqrt n\{T_K(\widehat\theta)-T_K(\theta)\}
\Rightarrow N\{0,J_K(\theta)\Sigma J_K(\theta)^\top\}.
\]
When the resulting variance vanishes, a second-order expansion is required.
\end{proposition}
\begin{proof}
By Proposition~\ref{prop:app-clt}, $\sqrt n(\widehat\theta-\theta)\Rightarrow N(0,\Sigma)$ in a finite-dimensional parameterization of the probability simplex and cell quantiles. Differentiability gives
\[
T_K(\widehat\theta)-T_K(\theta)=J_K(\theta)(\widehat\theta-\theta)+o_{\Pbb}(n^{-1/2}).
\]
The stated limit follows from Slutsky's theorem. If $J_K(\theta)\Sigma J_K(\theta)^\top=0$, the linear term is degenerate and the leading behavior is second order.
\end{proof}

\subsection{Quadratic limit under exact joint parity}
At $Z\indep\bS$, all population full-cell quantile vectors are equal. The first derivative of the quadratic disparity therefore vanishes.

\begin{proposition}[Quadratic null limit]\label{prop:app-null}
Under $Z\indep\bS$ and the density conditions above, let $G_s^q$ be the Gaussian cell-quantile limits and $\overline G=\sum_sp_sG_s^q$. Then
\[
n\widehat\Ujoint_K\Rightarrow
\sum_sp_s\|G_s^q-\overline G\|_K^2.
\]
Thus the exact-null limit is a weighted quadratic form in Gaussian variables rather than a centered normal law.
\end{proposition}

\begin{proof}
Under the null, $q_s=q_0$ for every cell. Let $\widehat{\bar q}=\sum_s\widehat p_s\widehat q_s$. Jointly,
\[
\sqrt n(\widehat q_s-q_0)\Rightarrow G_s^q,
\qquad
\sqrt n(\widehat{\bar q}-q_0)\Rightarrow\overline G.
\]
Hence $\sqrt n(\widehat q_s-\widehat{\bar q})\Rightarrow G_s^q-\overline G$. Substitution into
\[
\widehat\Ujoint_K=\sum_s\widehat p_s\|\widehat q_s-\widehat{\bar q}\|_K^2
\]
and Slutsky's theorem give the result. First-order fluctuations of the cell proportions multiply the common $q_0$ and cancel because the empirical weights sum to one.
\end{proof}

This degeneracy motivates the permutation calibration used in the main text. If a statistic or an ordering is selected adaptively, the complete selection rule must be rerun inside each permutation replicate for the permutation p-value to calibrate the data-dependent procedure rather than only the selected statistic.

\subsection{Rare-cell shrinkage}
A transparent stabilization is to shrink each empirical cell quantile vector toward the pooled empirical quantile vector $\widehat q_0$:
\[
\widetilde q_s=(1-\lambda_s)\widehat q_s+\lambda_s\widehat q_0,
\qquad
\lambda_s=\frac{\tau_n}{n_s+\tau_n}.
\]

\begin{proposition}[Asymptotically negligible shrinkage]\label{prop:app-shrink}
Assume the number of population cells is fixed and $p_s>0$ for each retained cell. If $\tau_n=o(n)$, then
\[
\max_s\|\widetilde q_s-\widehat q_s\|_K=o_{\Pbb}(1).
\]
If moreover $\tau_n=o(\sqrt n)$, then shrinkage leaves the first-order limit of every differentiable finite-grid diagnostic unchanged.
\end{proposition}

\begin{proof}
Since $n_s/n\to p_s$ uniformly over finitely many cells, $\max_s\lambda_s=O_{\Pbb}(\tau_n/n)$. The empirical and pooled fixed-grid quantile vectors are $O_{\Pbb}(1)$, so
\[
\max_s\|\widetilde q_s-\widehat q_s\|_K
=O_{\Pbb}(\tau_n/n).
\]
The first conclusion follows from $\tau_n=o(n)$; multiplying by $\sqrt n$ proves the second when $\tau_n=o(\sqrt n)$.
\end{proof}

\section{Algorithms and computational complexity}
\label{app:algorithms}
\label{app:computation}

Let $m$ be the number of retained nonempty full cells. In one dimension, Wasserstein barycenters are just weighted averages of quantile vectors, so the computational bottleneck is combinatorial rather than transport-theoretic.

\begin{table}[h]
\centering
\small
\begin{tabular}{ll}
\toprule
Task & Leading cost \\
\midrule
Sort scores within full cells & $O(\sum_s n_s\log n_s)$ \\
Extract $K$ grid quantiles & $O(mK)$ after sorting \\
Compute all coalition values & $O(2^d mK)$ \\
Exact Shapley from cached coalitions & $O(d2^d)$ \\
All order profiles & $O(d!\,d)$ after caching \\
Monte Carlo Shapley, $B$ permutations
& $O(BdmK)$ without full coalition caching \\
Bootstrap/permutation, $R$ replicates & $R$ times chosen audit cost \\
\bottomrule
\end{tabular}
\caption{Leading computational costs of the finite-grid implementation. Exact coalition enumeration is intended for moderate $d$; permutation Shapley is the natural approximation for larger sensitive profiles.\\
After caching all coalition values, the marginal lookup cost is $O(Bd)$,
but this requires the preceding $O(2^d mK)$ preprocessing step.}
\label{tab:app-complexity}
\end{table}

The finite-grid implementation has modest memory requirements. Storing the $m$ retained cellwise quantile vectors costs $O(mK)$ memory, while caching all coalition values for exact Shapley computation requires an additional $O(2^d)$ scalars. Thus exact enumeration can be implemented with $O(mK+2^d)$ memory; permutation-based Shapley computation only requires the cellwise quantiles and the values along sampled paths, reducing the additional working memory to $O(d)$. The computational and statistical bottlenecks are distinct: the former is driven by the $2^d$ coalition lattice, whereas the latter is driven by the number and occupancy of the full sensitive cells. Permutation sampling alleviates the first bottleneck but cannot compensate for sparse or empty intersections.
For a random ordering $\pi$, let $P_j^\pi$ denote the set of attributes preceding $j$ and write
\[
H_j^\pi
:=
H_{j\mid P_j^\pi}
=
\mathcal V(P_j^\pi\cup\{j\})-\mathcal V(P_j^\pi).
\]
The permutation representation of the Shapley value is then
\[
\phi_j=\E_\pi H_j^\pi.
\]
Given $B$ i.i.d.\ uniformly sampled permutations $\pi_1,\ldots,\pi_B$, define
\[
\widetilde\phi_j^{(B)}
=
\frac1B\sum_{b=1}^B H_j^{\pi_b}.
\]

\begin{proposition}[Exact efficiency and Monte Carlo error]
\label{prop:app-mc}
For every $B\ge1$,
\[
\sum_{j=1}^d \widetilde\phi_j^{(B)}
=
\mathcal V([d])
\]
exactly. Moreover,
\[
\E_\pi\!\left\{\widetilde\phi_j^{(B)}\right\}
=
\phi_j,
\qquad
\Var_\pi\!\left\{\widetilde\phi_j^{(B)}\right\}
=
\frac1B\Var_\pi(H_j^\pi).
\]
If $0\le H_j^\pi\le R_j$ for every permutation $\pi$, then
\[
\Pbb_\pi\!\left(
\left|\widetilde\phi_j^{(B)}-\phi_j\right|>\varepsilon
\right)
\le
2\exp\!\left(
-\frac{2B\varepsilon^2}{R_j^2}
\right).
\]
\end{proposition}

\begin{proof}
For every sampled permutation $\pi_b$, the $d$ barycentric path increments telescope:
\[
\sum_{j=1}^d H_j^{\pi_b}
=
\mathcal V([d])-\mathcal V(\varnothing)
=
\mathcal V([d]).
\]
Averaging this identity over $b=1,\ldots,B$ proves exact finite-$B$ efficiency. For a fixed $j$, the variables $H_j^{\pi_1},\ldots,H_j^{\pi_B}$ are i.i.d.\ under uniform permutation sampling, with mean $\E_\pi H_j^\pi=\phi_j$. Unbiasedness and the $1/B$ variance formula follow immediately. The final bound is Hoeffding's inequality applied to their sample mean.
\end{proof}

The exact-efficiency statement is algebraic rather than asymptotic: each sampled reveal path allocates the same total $\mathcal V([d])$. The analogous identity also holds for plug-in coalition values, with $\mathcal V([d])$ replaced by $\widehat{\mathcal V}_K([d])$.

To separate the sources of approximation, let $\phi_{j,K}$ denote the population Shapley value of the finite-grid game and $\widehat\phi_{j,K}$ its exact plug-in counterpart. Then
\[
\widehat\phi_j^{(B)}-\phi_j
=
\underbrace{
\bigl(\widehat\phi_j^{(B)}-\widehat\phi_{j,K}\bigr)
}_{\text{permutation sampling}}
+
\underbrace{
\bigl(\widehat\phi_{j,K}-\phi_{j,K}\bigr)
}_{\text{statistical estimation}}
+
\underbrace{
\bigl(\phi_{j,K}-\phi_j\bigr)
}_{\text{grid approximation}}.
\]
Away from first-order degeneracy, the results of
Appendix~\ref{app:statistics} therefore give
\[
\widehat\phi_j^{(B)}-\phi_j
=
O_{\Pbb}(B^{-1/2})
+
O_{\Pbb}(n^{-1/2})
+
O(K^{-\alpha}).
\]
This is the standard permutation representation of Shapley values \citep{castro2009polynomial,maleki2013bounding}.

\section{Extended numerical evidence}
\label{app:numerics}
The simulation parameters below correspond to the scripts used to generate the reported experiments. The code uses R's type-8 empirical quantiles throughout categorical Wasserstein experiments; the paper-run configuration uses 99 random permutations for null calibration and 99 pairs-bootstrap resamples inside each Gaussian Monte Carlo replication.

\subsection{Marginal-versus-joint power experiment}
The main-text power experiment uses three independent Rademacher sensitive variables and $\varepsilon\sim N(0,1)$. We compare
\[
Z=0.15S_1+\varepsilon
\qquad\text{and}\qquad
Z=0.15S_1S_2S_3+\varepsilon.
\]
For each $n\in\{200,400,800,1600\}$ we generated 100 independent datasets. The joint statistic is the fixed-grid plug-in $\widehat U_{\rm joint}$ and the marginal statistic is $\sum_{j=1}^3\widehat U_j$, both using the nine-point grid $u=.1,.2,\ldots,.9$. Each statistic is calibrated separately with 99 score permutations and the corrected randomization p-value $(1+r)/(100)$. The resulting rejection proportions are:
\begin{center}
\begin{tabular}{lrrrr}
\toprule
& $n=200$ & $400$ & $800$ & $1600$\\
\midrule
Main effect, joint & .25 & .59 & .89 & 1.00\\
Main effect, marginal & .37 & .80 & .96 & 1.00\\
Pure interaction, joint & .27 & .51 & .84 & 1.00\\
Pure interaction, marginal & .04 & .02 & .04 & .05\\
\bottomrule
\end{tabular}
\end{center}
The experiment is intentionally simple. Its purpose is not to rank the tests universally, but to show the finite-sample cost of estimating eight full-cell distributions when the alternative is only first-order, and the complementary impossibility of detecting a pure three-way alternative with coordinate-wise marginal audits.

\subsection{Exact null: finite-sample bias and permutation size}
The null experiment uses independent Rademacher sensitive attributes and $Z\sim N(0,1)$ independently of the sensitive profile, so $\Ujoint=0$ in the population. The paper run uses $d\in\{2,3,4,5\}$, $n\in\{500,1000,5000,10000\}$, $K\in\{25,50,100\}$, and pseudo-counts $\tau\in\{0,20\}$. For each $(d,n,K,\tau)$ configuration we generate 100 independent datasets and calibrate the same statistic with 99 score permutations. The grid is $u_k=k/(K+1)$ and empirical cell quantiles use R's type-8 interpolation. The minimum retained cell size is one in this simulation, so the experiment isolates sampling sparsity rather than threshold selection.

Even at the exact null the raw empirical statistic is positive because independent empirical cell quantiles fluctuate around the same population quantile curve. The relevant effective sample size is the average number of observations per full cell, $n/2^d$: the positive bias decreases with this quantity, and $\tau=20$ reduces it most visibly in sparse-cell regimes. Despite the positive raw statistic, the corrected permutation test
\[
\widehat p=\frac{1+\sum_{b=1}^{99}\mathbf 1\{T^{(b)}\ge T^{\rm obs}\}}{100}
\]
remains close to the nominal $5\%$ level across the explored dimensions and cell sizes. These experiments motivate reporting cell sizes and using permutation rather than first-order Wald calibration at the exact null.

\subsection{Pure highest-order interaction}
Consider independent Rademacher sensitive attributes and
\[
Q_{\bS}(u)=Q_0(u)+a\prod_{j=1}^dS_j.
\]
Then $\Ujoint=a^2$, every strict lower-order coalition value is zero, every path decomposition assigns the population interaction to the last entering variable, and Shapley divides it equally: $\phi_j=a^2/d$.
The paper run sets $a=1$, adds independent $N(0,0.5^2)$ noise, uses $d\in\{3,4,5\}$ and $n\in\{500,1000,5000,10000\}$, and averages over 200 Monte Carlo datasets per configuration. Coalition values use the 99-point grid $u=.01,.02,\ldots,.99$ with type-8 empirical quantiles.

\begin{table}[h]
\centering
\scriptsize
\begin{tabular}{rrrrrrrrr}
\toprule
$d$ & $n$ & True $U$ & Mean $\widehat U$ & Bias & RMSE & max strict $\widehat V$ & $\phi$ RMSE & last share \\
\midrule
3&500&1.000&1.006&0.006&0.048&0.0142&0.0135&0.990\\
3&1000&1.000&1.003&0.003&0.030&0.0070&0.0084&0.995\\
3&5000&1.000&1.002&0.002&0.015&0.0013&0.0039&0.999\\
3&10000&1.000&0.999&-0.001&0.010&0.0006&0.0027&1.000\\
4&500&1.000&1.017&0.017&0.051&0.0297&0.0111&0.977\\
4&1000&1.000&1.009&0.009&0.033&0.0149&0.0069&0.989\\
4&5000&1.000&1.000&0.000&0.014&0.0030&0.0028&0.998\\
4&10000&1.000&1.000&0.000&0.010&0.0015&0.0021&0.999\\
5&500&1.000&1.027&0.027&0.053&0.0573&0.0101&0.955\\
5&1000&1.000&1.018&0.018&0.035&0.0283&0.0061&0.977\\
5&5000&1.000&1.003&0.003&0.015&0.0060&0.0025&0.995\\
5&10000&1.000&1.003&0.003&0.009&0.0031&0.0016&0.998\\
\bottomrule
\end{tabular}
\caption{Pure highest-order interaction. The plug-in estimator recovers the joint disparity, strict lower-order coalition values vanish with $n$, and the last martingale increment approaches the full joint signal. The corresponding Shapley values sum to $\widehat U$ to numerical precision.}
\label{tab:app-pure}
\end{table}

\subsection{Gaussian suppression: exact calculations}
Table~\ref{tab:app-gauss-calc} illustrates the formulas from Section~\ref{app:theory}. The balanced case makes the conditional-to-marginal ratio explode monotonically; the stronger-suppression case has exact marginal cancellation at $\rho=2/3$ even though the conditional increment remains $5/9$.

\begin{table}[h]
\centering
\small
\begin{tabular}{c|rrr|rrr}
\toprule
&\multicolumn{3}{c|}{$(\beta_1,\beta_2)=(-1,1)$}&\multicolumn{3}{c}{$(\beta_1,\beta_2)=(-1.5,1)$}\\
$\rho$ & $\mathcal V(\{2\})$ & $H_{2\mid1}$ & ratio & $\mathcal V(\{2\})$ & $H_{2\mid1}$ & ratio\\
\midrule
0.00&1.0000&1.0000&1.0&1.0000&1.0000&1.0\\
0.50&0.2500&0.7500&3.0&0.0625&0.7500&12.0\\
$2/3$&0.1111&0.5556&5.0&0.0000&0.5556&$\infty$\\
0.80&0.0400&0.3600&9.0&0.0400&0.3600&9.0\\
0.90&0.0100&0.1900&19.0&0.1225&0.1900&1.55\\
0.95&0.0025&0.0975&39.0&0.1806&0.0975&0.54\\
\bottomrule
\end{tabular}
\caption{Closed-form suppression calculations. The ratio is $H_{2\mid1}/\mathcal V(\{2\})$. In the stronger-suppression case the marginal signal cancels exactly at $\rho=2/3$ and then changes sign before squaring, so the ratio is not monotone beyond the cancellation point.}
\label{tab:app-gauss-calc}
\end{table}

The finite-sample experiment uses $\rho\in\{0,.25,.50,.75,.90\}$, $n\in\{500,1000,5000,10000\}$, two coefficient cases $(\beta_1,\beta_2)=(-1,1)$ and $(-1.5,1)$, and 200 Monte Carlo datasets per configuration. Within every Monte Carlo dataset, uncertainty for $\mathcal V(\{2\})$ and $H_{2\mid1}$ is computed by a nonparametric pairs bootstrap with 99 resamples: the $n$ observed rows are sampled with replacement, the statistic is recomputed, and the reported $95\%$ interval is the percentile interval between the empirical $2.5\%$ and $97.5\%$ bootstrap quantiles. The plug-in estimators track both population curves across the correlation grid. Coverage is broadly close to nominal away from near-degenerate configurations, while some $n=500$ cases under-cover for conditional increments; this is consistent with the quadratic/near-degenerate warning in Section~\ref{app:statistics}. The bootstrap intervals here are descriptive finite-sample diagnostics rather than a claim of uniform validity at degeneracy.

\subsection{Computation time}
\begin{table}[h]
\centering
\small
\begin{tabular}{lrrrrr}
\toprule
Experiment & Configurations & Mean sec. & Median sec. & 90\% sec. & Max sec.\\
\midrule
Null & 96 & 0.01 & 0.01 & 0.02 & 0.02\\
Pure interaction & 12 & 0.01 & 0.01 & 0.02 & 0.02\\
Gaussian suppression & 200 & 0.24 & 0.19 & 0.50 & 0.50\\
\bottomrule
\end{tabular}
\caption{Average computation time per Monte Carlo replicate in the long-form simulation study. Times are machine-dependent but confirm feasibility for moderate sensitive dimension and finite grids.}
\label{tab:app-runtime}
\end{table}

\section{Extended \texttt{ACSIncome} results}
\label{app:acs}
The empirical audit is performed on held-out regularized-logistic scores, standardized within state. Age is discretized into four groups, the grid is $u=.01,.02,\ldots,.99$, cells with fewer than 20 held-out observations are removed, and 999 score permutations across intact sensitive profiles calibrate the joint null. Stratified bootstrap intervals resample scores within full sensitive cells.

\begin{table}[h]
\centering
\small
\begin{tabular}{lrrrrl}
\toprule
State & $n$ & cells & min cell & median cell & $\Ujoint$ [bootstrap interval]\\
\midrule
CA&9000&32&22&219.0&0.0173 [0.0163, 0.0192]\\
NY&9000&32&26&135.5&0.0116 [0.0109, 0.0129]\\
IL&8938&28&42&91.0&0.0142 [0.0134, 0.0156]\\
GA&8969&29&20&155.0&0.0122 [0.0114, 0.0135]\\
TX&9000&32&21&96.0&0.0120 [0.0113, 0.0132]\\
OH&8929&26&21&100.0&0.0132 [0.0125, 0.0145]\\
PA&8962&28&22&58.0&0.0133 [0.0126, 0.0146]\\
MI&8956&28&29&67.5&0.0129 [0.0123, 0.0142]\\
NC&8970&29&24&143.0&0.0098 [0.0093, 0.0109]\\
FL&9000&32&20&72.5&0.0092 [0.0087, 0.0102]\\
\bottomrule
\end{tabular}
\caption{\texttt{ACSIncome}: held-out sample sizes, retained full cells, cell-size diagnostics, and joint Wasserstein disparity. In every state, the 999-permutation p-value is $\le0.001$; all 99\% permutation thresholds are below $0.0011$.}
\label{tab:app-acs-main}
\end{table}

\begin{table}[h]
\centering
\small
\begin{tabular}{lrrrrr}
\toprule
State & $\Uworst/\Ujoint$ & order sens. (\%) & Sex (\%) & Race (\%) & Age (\%)\\
\midrule
CA&4.42&0.18&2.0&22.1&75.8\\
NY&7.38&0.76&3.2&17.1&79.7\\
IL&5.39&0.21&4.6&15.7&79.8\\
GA&4.82&0.90&5.9&14.3&79.8\\
TX&5.87&0.93&5.2&13.4&81.4\\
OH&5.23&0.36&3.5&10.2&86.4\\
PA&5.53&0.24&3.4&8.2&88.5\\
MI&7.99&0.48&5.0&11.3&83.6\\
NC&5.83&1.15&3.9&19.0&77.1\\
FL&6.68&1.11&5.0&11.1&83.9\\
\bottomrule
\end{tabular}
\caption{\texttt{ACSIncome}: concentration, relative sequential order sensitivity, and Shapley shares. The three Shapley columns sum to $100\%$ up to rounding and allocate the fixed $\Ujoint$ descriptively, not causally.}
\label{tab:app-acs-shapley}
\end{table}

The observed joint disparities are roughly one to two orders of magnitude above the permutation-null scale: null means are around $3\times10^{-4}$--$5\times10^{-4}$ and all 99\% null quantiles are below $0.0011$. The ratio $\Uworst/\Ujoint$ ranges from $4.42$ to $7.99$, whereas relative sequential order sensitivity is below $1.2\%$ in all states. Thus the most discrepant retained intersection can be much farther from the barycentric reference than the average cell even when the aggregate sequential audit is almost insensitive to order. The Shapley allocation is stable in qualitative ranking: age dominates, race is second, and sex is smallest in every reported state.

\section{Practical reporting checklist}
For a reproducible multi-sensitive audit, we recommend reporting: (i) the sensitive coordinates and how continuous coordinates were discretized; (ii) the number of possible and retained full cells, minimum and median retained cell size, and any merging/exclusion rule; (iii) the quantile grid and any shrinkage pseudo-count; (iv) $\Ujoint$ together with a calibrated permutation p-value for the exact joint null; (v) a concentration diagnostic such as $\Uworst/\Ujoint$ when extreme intersections matter; (vi) Shapley shares together with the raw $\Ujoint$ they decompose; and (vii) an explicit statement that the allocation is descriptive unless additional causal assumptions are imposed.

\end{document}